\pdfoutput=1
\documentclass[pmlr,twocolumn,10pt]{jmlr}

\newif\ifhtmlbuild
\ifdefined\mlhtrack\else\htmlbuildtrue\fi
\ifhtmlbuild
  \providecommand{\mlhtrack}[1]{}
  \providecommand{\mlhtrackname}{}
  \providecommand{\jmlrvolume}[1]{}
  \providecommand{\jmlryear}[1]{}
  \providecommand{\jmlrworkshop}[1]{}
  \providecommand{\jmlrproceedings}[2]{}
  \expandafter\newif\csname ifmlhneedspmlr\endcsname
  \expandafter\newif\csname ifmlhneedsstatements\endcsname
  \csname mlhneedsstatementstrue\endcsname
\fi

\mlhtrack{proceedings}
\newif\iffinal
\finaltrue

\iffinal
  \ifmlhneedspmlr
    \jmlrvolume{346}
    \jmlryear{2026}
  \fi
  \jmlrworkshop{Machine Learning for Health (ML4H) 2026}
\else
  \jmlrproceedings{}{Submitted to ML4H 2026: \mlhtrackname}
  \jmlrworkshop{Machine Learning for Health (ML4H) 2026}
\fi

\usepackage{booktabs}
\usepackage{siunitx}
\usepackage{fontawesome5}
\usepackage[switch]{lineno}

\hypersetup{
  pdftitle={SLICES: A Benchmark and Evaluation Framework for Self-Supervised Learning on Sparse ICU Time Series},
  pdfauthor={Hannes Ill, Dimitrios Proios, Leo Anthony Celi, Peter J. Sch\"uffler},
  pdfsubject={ML4H 2026 Proceedings},
  pdfkeywords={self-supervised learning, clinical time series, intensive care, benchmark, missing data}
}

\newcommand{\SLICES}{SLICES}
\newcommand{\MIMIC}{MIMIC-IV}
\newcommand{\EICU}{eICU-CRD}
\newcommand{\Combined}{Combined}
\newcommand{\MAE}{MAE}
\newcommand{\JEPA}{JEPA}

\defcitealias{kdigo2012aki}{KDIGO AKI Work Group, 2012}

\newcommand{\releaseurl}{https://github.com/hannesill/SLICES/tree/ml4h-2026}

\title[SLICES: Benchmarking SSL on Sparse ICU Time Series]{SLICES: A Benchmark and Evaluation Framework for Self-Supervised Learning on Sparse ICU Time Series}
\ifhtmlbuild
\author{Hannes Ill\\
  {Laboratory for Computational Physiology, Massachusetts Institute of Technology, USA}\\
  {Institute of Pathology, Technical University of Munich, TUM School of Medicine and Health, Munich, Germany}\\
  \texttt{illh@mit.edu}
  \and Dimitrios Proios\\
  {Department of Radiology and Medical Informatics, Faculty of Medicine, University of Geneva, Geneva, Switzerland}\\
  \texttt{dimitrios.proios@unige.ch}
  \and Leo Anthony Celi\\
  {Laboratory for Computational Physiology, Massachusetts Institute of Technology, USA}\\
  {Division of Pulmonary, Critical Care and Sleep Medicine, Beth Israel Deaconess Medical Center, USA}\\
  \texttt{lceli@mit.edu}
  \and Peter J. Sch\"uffler\\
  {Institute of Pathology, Technical University of Munich, TUM School of Medicine and Health, Munich, Germany}\\
  {Munich Center for Machine Learning (MCML), Munich, Germany}\\
  {Munich Data Science Institute (MDSI), Munich, Germany}\\
  \texttt{peter.schueffler@tum.de}}
\else
\author{%
  \Name{Hannes Ill} \Email{illh@mit.edu}\\
  \addr Laboratory for Computational Physiology, Massachusetts Institute of Technology, USA\\
  \addr Institute of Pathology, Technical University of Munich, TUM School of Medicine and Health, Munich, Germany
  \AND
  \Name{Dimitrios Proios} \Email{dimitrios.proios@unige.ch}\\
  \addr Department of Radiology and Medical Informatics, Faculty of Medicine, University of Geneva, Geneva, Switzerland
  \AND
  \Name{Leo Anthony Celi} \Email{lceli@mit.edu}\\
  \addr Laboratory for Computational Physiology, Massachusetts Institute of Technology, USA\\
  \addr Division of Pulmonary, Critical Care and Sleep Medicine, Beth Israel Deaconess Medical Center, USA
  \AND
  \Name{Peter J. {Sch\"uffler}} \Email{peter.schueffler@tum.de}\\
  \addr Institute of Pathology, Technical University of Munich, TUM School of Medicine and Health, Munich, Germany\\
  \addr Munich Center for Machine Learning (MCML), Munich, Germany\\
  \addr Munich Data Science Institute (MDSI), Munich, Germany
}
\fi

\iffinal\else\linenumbers\fi

\begin{document}
\maketitle

\begin{abstract}
Self-supervised learning (SSL) results for intensive care unit (ICU) time series are hard to compare when data processing, encoders, and adaptation vary together. We present \SLICES{}, a benchmark and evaluation framework for comparing and extending SSL methods under a shared clinical setup. It fixes 24-hour value--observation-mask inputs, a Transformer encoder, patient-level splits, four endpoints, two adaptation protocols, and five matched seeds across \MIMIC{}, \EICU{}, and their pooled union. Masked autoencoding (\MAE{}), joint-embedding prediction (\JEPA{}), and contrastive learning provide reference implementations. Among these objectives, \MAE{} leads mean AUPRC in all nine frozen classification comparisons, also under converged probes, and in seven of nine fine-tuning comparisons. Fine-tuned \MAE{} leads hospital mortality at full labels and scores higher than a supervised Transformer of the same architecture at every label fraction, and SSL fine-tuning gives the lowest length-of-stay error. Supervised baselines remain strong elsewhere: GRU-D leads 24-hour mortality and XGBoost leads acute kidney injury in all three regimes. Rankings depend on the adaptation protocol, the comparator set, and whether the probe loss matches the metric. A two-seed \MIMIC{} rerun with the maintained code stays within two across-seed SDs of the reference in all 48 downstream runs. The code repository supports new objectives, shared downstream evaluation, and regeneration of the reference results, so new SSL methods can be compared with these references under the same conditions. Generalization to unseen hospitals is not tested here and is the natural next extension.
\end{abstract}


\ifmlhneedsstatements
\paragraph*{Data and Code Availability}
We use deidentified \MIMIC{} v3.1 and \EICU{} v2.0 data from PhysioNet under credentialed access \citep{physionet2024mimiciv31,physionet2019eicu20}. Code, run-level aggregate results, post-review diagnostics, and executable audits are available in the code repository at \href{\releaseurl}{\texttt{github.com/hannesill/SLICES}} (tag \texttt{ml4h-2026}). The repository excludes patient records, split identities, credentials, and trained patient-data weights. Aggregate auditing needs no data access. Training reproduction requires independent credentialing, training, and a data-use agreement.

\paragraph*{Institutional Review Board (IRB)}
This secondary analysis of deidentified, credentialed-access data under the PhysioNet data-use agreements does not require IRB approval.
\fi

\section{Introduction}
\label{sec:introduction}

\begin{figure*}[!t]
\floatconts{fig:thesis-overview}
{\caption{Benchmark overview. Cohorts, hourly inputs, encoder architecture, patient splits, tasks, and evaluation protocols are fixed. SSL implementations vary in their masks and auxiliary modules. Equal epochs do not imply equal compute. Baselines provide context, and TS2Vec-style illustrates extension. LOS uses mean absolute error in days.}}
{\includegraphics[width=.92\textwidth]{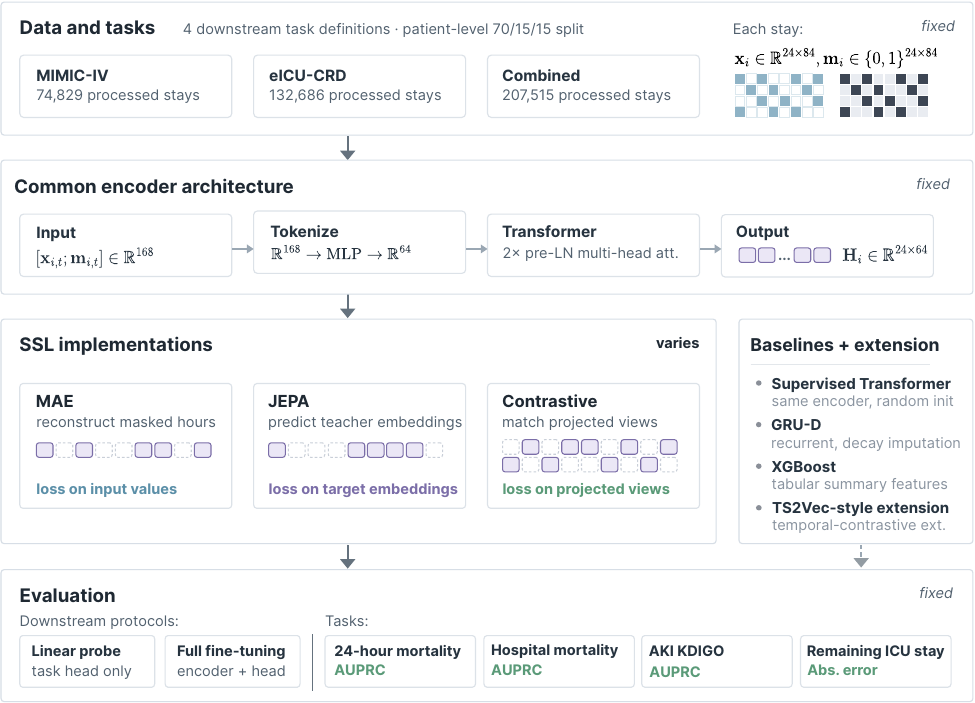}}
\end{figure*}

ICU records are sparse because measurements are taken at different frequencies and often in response to a patient's evolving condition. The absence and timing of a measurement can therefore carry information about both physiology and care processes \citep{tan2023informativemissingness}. This makes ICU time series a relevant test bed for SSL. Unlabeled measurements are abundant, but the measurement process complicates the choice of learning signal.

Prior clinical SSL studies use reconstruction, forecasting, contrastive learning, or combined pretext tasks \citep{tipirneni2022strats,labach2023duett,yeche2021ncl,oufattole2024ebcl}. Their differences are hard to attribute when cohorts, representations, encoders, augmentations, and evaluation also vary. A systematic review of structured-EHR foundation models found similar heterogeneity in features, temporal representation, SSL objectives, and adaptation, with compute often underreported \citep{guo2026ehrfoundationreview}. Frozen probes test whether task information is linearly accessible. Fine-tuning tests whether a pretrained initialization remains useful after label-driven adaptation. Either alone can hide a protocol-dependent ranking.

\SLICES{} makes the evaluation contract reusable (Figure~\ref{fig:thesis-overview}). A new objective consumes the same value--mask tensors, returns a loss through a common interface, and exports the same encoder for downstream evaluation. A machine-readable specification and result validators make changes to the comparison explicit. The controlled unit is the complete implementation: masking geometry, auxiliary networks, encoder passes, and pretraining parameter counts remain objective specific.

Our contributions are:
\begin{enumerate}
  \item A benchmark framework: a fixed sparse-ICU specification, an objective/encoder-transfer interface with MAE, JEPA, and contrastive reference implementations, result validators, and a documented extension workflow. It fixes one representation, hourly values with observation masks.
  \item A first study run with the framework: 2,455 training runs across two databases and their union, with five matched seeds. GRU-D and XGBoost lead six of twelve full-label cells, fine-tuned SSL leads hospital mortality narrowly and LOS, and \MAE{} leads all frozen SSL classification comparisons. Rankings depend on the adaptation protocol, label fraction, comparator set, and probe loss.
  \item A code repository with the historical training implementation, framework code, checksummed aggregate results, and tools to regenerate results and check comparability without patient-data access.
\end{enumerate}
 
\section{Related Work}
\label{sec:related}

\paragraph{ICU benchmarks.}
The MIMIC-III benchmark formalized common ICU prediction tasks and evaluation procedures \citep{harutyunyan2019multitask}. YAIB provided a modular multi-center clinical framework over several ICU databases using harmonized preprocessing \citep{van_de_water2024yaib}. \SLICES{} builds on explicit cohorts, labels, and leakage controls with a setup designed for SSL comparisons. It holds the clinical inputs, encoder, and downstream evaluation fixed across objectives, reducing variation from these sources.

\paragraph{Clinical time-series SSL.}
STraTS represents sparse observations as time--variable--value triplets and uses forecasting-style pretraining \citep{tipirneni2022strats}. DuETT uses dual event-time attention and self-supervised prediction tasks \citep{labach2023duett}. Neighborhood and event-based contrastive methods use clinical structure to define positive pairs or anchors \citep{yeche2021ncl,oufattole2024ebcl}. Sequential multi-dimensional SSL evaluates contrastive and non-contrastive losses at multiple levels of clinical trajectories \citep{raghu2023smdssl}. SMART combines missing-aware temporal and variable attention with latent reconstruction \citep{yu2024smart}. AID-MAE is a recent masked-modeling approach for incomplete EHR time series \citep{xiang2026aidmae}. These systems motivate the objectives studied here but change either the representation, architecture, positive-pair rule, or task contract.

\paragraph{Closest comparative work.}
TS-JEPA compares JEPA, masked autoencoding, autoregression, and TS2Vec-style contrastive pretraining with a common Transformer, on sensor and forecasting datasets, with frozen-head evaluation and a low-label study \citep{ennadir2024tsjepa}. It is the closest objective-level predecessor. \SLICES{} differs in its missingness-preserving ICU interface, patient-level clinical splits, paired probe and fine-tuning evaluation, and matched \MIMIC{}--\EICU{} regimes. Clin-JEPA uses text-serialized hourly states and an LLM-based encoder, so it changes the model and input contract rather than testing a shared small-encoder interface \citep{yang2026clinjepa}. General MAE, I-JEPA, SimCLR, and TS2Vec formulations provide the objective context \citep{he2022mae,assran2023ijepa,chen2020simclr,yue2022ts2vec}.

Scores from these studies are not directly comparable with SLICES because cohorts, representations, and tasks differ.

\section{Benchmark Design}
\label{sec:design}

\subsection{Cohorts and Hourly Representation}

We use \MIMIC{} v3.1 and \EICU{} v2.0 \citep{johnson2023mimiciv,pollard2018eicu}. Common extraction and unit harmonization are based on RICU \citep{bennett2023ricu}. Each sample is one ICU stay with valid patient and stay metadata, ICU duration of at least 24 hours, and at least one observed time-series value in the first 24 hours. The processed cohorts contain 74,829 \MIMIC{} stays and 132,686 \EICU{} stays. We form their 207,515-stay union, \emph{Combined}, after schema compatibility checks and source-specific identifier namespacing. \MIMIC{}, \EICU{}, and \Combined{} are the three regimes; \Combined{} is pooled, not a separate cohort.

Each stay is represented as normalized values $x\in\mathbb{R}^{24\times84}$ and an observation mask $m\in\{0,1\}^{24\times84}$. Feature means and standard deviations are fitted on the training partition only. Missing cells are zero filled after normalization while $m$ preserves whether a value was observed. RICU aggregates repeated measurements within an hour by concept-specific rules, by default the median for numeric values and ``any'' for logical indicators, with exceptions such as summed urine output. The mask records whether at least one value fell in an hour, not how many, so within-hour sampling intensity is lost. The 84 retained variables are shared across regimes. Only 10.7\% to 15.3\% of tensor cells are observed, although most stays have at least one measurement in nearly every hour.

Patients are assigned to 70/15/15 train, validation, and test partitions for each seed. All stays for one patient remain in one partition. Task-specific missing labels are removed only after this assignment, and the same split is used for pretraining and downstream evaluation. Input normalization uses training data only. Direct length-of-stay variables, future-duration helpers, and derived features that expose downstream outcomes are excluded from the input.

\subsection{Endpoints and Horizons}

All tasks use the first 24 ICU hours as input. The endpoints follow task families of established ICU benchmarks \citep{harutyunyan2019multitask,van_de_water2024yaib}. They differ in prevalence, horizon, label construction, and target type, which can change an objective ranking. Classification uses area under the precision--recall curve (AUPRC), computed as average precision, because prevalence is low and endpoint dependent \citep{saito2015precisionrecall}. Mean test prevalence spans 1.6\%--20.3\% (\tableref{tab:test-support}). Twenty-four-hour mortality predicts death in hours 24--48. Hospital mortality predicts in-hospital death after hour 24. Deaths within the input window receive null labels. Creatinine-defined acute kidney injury (AKI) uses the minimum in hours 0--24 as baseline and seeks KDIGO stage~$\geq1$ evidence in hours 24--48: a later value at least 1.5 times baseline or at least 0.3 mg/dL above an earlier value within 48 hours \citepalias{kdigo2012aki}. Urine output is excluded. A missing baseline or forward measurement yields a null label. Remaining ICU length of stay (LOS) is ICU duration minus one day, clipped at zero. We retain this continuous target rather than impose a binary threshold. Absolute error keeps prediction misses interpretable in days.

\begin{figure*}[t]
\floatconts{fig:ssl-objectives}
{\caption{Core objective geometries. MAE reconstructs observed values at masked hours. JEPA's online encoder receives visible valid hours. Its EMA target encoder receives the full unmasked valid sequence. JEPA predicts layer-normalized, stop-gradient targets, with loss only on masked valid hours. Contrastive learning aligns projected pooled views. Auxiliary modules are discarded before downstream transfer.}}
{\includegraphics[width=\textwidth]{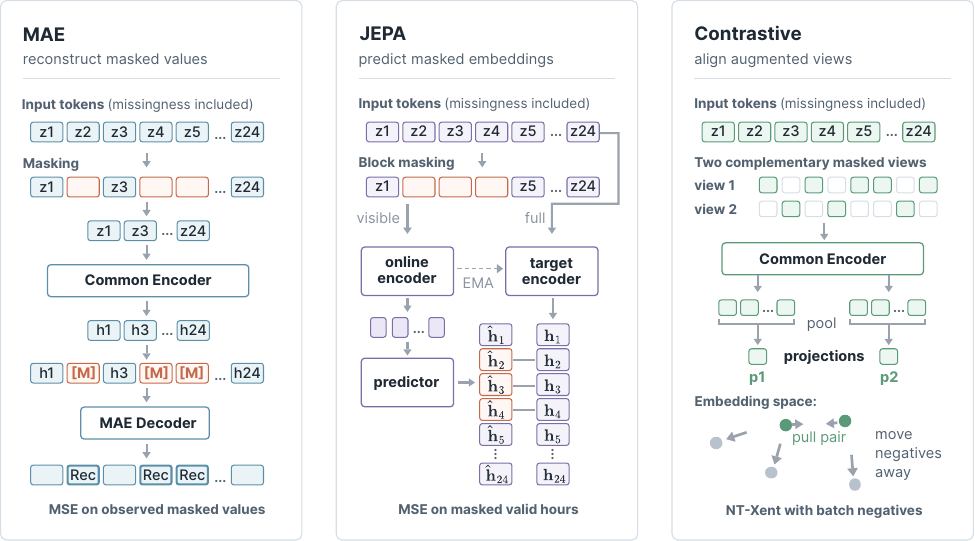}}
\end{figure*}

\subsection{Shared Encoder and Objective-Specific Components}

Appendix Figure~\ref{fig:encoder-tokenization} shows the shared interface. Each hour is one continuous attention token: 84 normalized values and 84 observation indicators are concatenated and projected from $\mathbb{R}^{168}$ to $\mathbb{R}^{64}$, without value quantization or vocabulary lookup. The shared small Transformer has sinusoidal time encoding, two pre-layer-normalized layers, four heads, feed-forward width 256, and dropout .1. Empty hours are excluded from attention and SSL losses. Downstream heads mean-pool valid outputs. The same 159,808-parameter encoder transfers for every SSL method.

Figure~\ref{fig:ssl-objectives} contrasts the learning signals. \MAE{} masks random non-empty hours, encodes visible tokens, and uses a two-layer Transformer decoder to reconstruct observed normalized values at masked hours. Missing cells do not enter the loss. \JEPA{} masks up to three contiguous blocks. An online encoder and predictor estimate layer-normalized embeddings from a stop-gradient exponential-moving-average target encoder. Contrastive learning makes two complementary masked views, encodes each separately, mean-pools visible hours, and applies an MLP projection with an NT-Xent instance loss at temperature .07. Only the common encoder transfers downstream.

The implementations preserve reconstruction, latent prediction, and view agreement as distinct learning signals while sharing inputs, transferred encoder, and evaluation. Forecasting objectives are out of scope. Objective-specific masks and auxiliary modules retain each method's learning structure. MAE has 269,460 trainable pretraining parameters, JEPA has 189,440 plus a non-trainable 159,808-parameter target copy, and contrastive learning has 193,408 with two encoder passes. These are comparisons of complete implementations, not causal loss-function effects. Appendix Tables~\ref{tab:pretraining} and~\ref{tab:objective-details} summarize the pretraining contract and modules.
 
\section{Experimental Setup}
\label{sec:setup}

\subsection{Using and Extending the Benchmark}
\label{sec:resource}

The code repository supports auditing the archived results (the aggregates of all 2,455 runs), regenerating tables and figures without retraining, reproducing the comparison with credentialed data access, and evaluating additional SSL objectives (Appendix Figure~\ref{fig:benchmark}). A machine-readable specification records clinical inputs, splits, tasks, encoder, training and checkpoint policies, and metrics. Comparability checks flag changed settings.

New objectives use the shared value--mask input and export an encoder for both frozen probing and fine-tuning. Pretraining retains hourly representations. Downstream evaluation averages valid hours into one stay-level vector because each endpoint assigns one target to a stay. Auxiliary pretraining networks are discarded. The TS2Vec-style extension demonstrates this path. A short run on artificial data checks the workflow without patient-data access. Its scores are not clinical evidence. Appendix~\ref{app:extension-workflow} gives implementation and validation details.

Historical source, current framework, archived results, and later diagnostics remain separately traceable. The framework comprises the specification, interface, validators, and extension workflow. The reference evaluation below is one study run through it.

\subsection{Reference Evaluation}

The study comprises 2,455 training runs: 145 pretraining and 2,310 downstream or baseline runs. It tests objective and protocol choice, reduced-label performance, cross-database adaptation, hyperparameter and encoder-capacity sensitivity, classical baselines, and the TS2Vec-style extension. The core comparison accounts for 45 pretraining runs and 420 evaluations.

\paragraph{Pretraining.}
All methods use 100 epochs, batch size 256, AdamW learning rate $10^{-3}$, weight decay .01, gradient clipping at 1.0, ten warmup epochs, and cosine decay to $10^{-6}$, without early stopping. Batches contain only training-partition stays, and validation loss is monitored. Equal epochs give an equal data-pass schedule within each regime, not equal FLOPs, wall time, or objective-specific module cost. Configurations, shared across datasets, were fixed before the final rerun but informed by earlier development passes, whose test metrics we saw while fixing label and input defects. Test partitions were never used for training, normalization, or checkpoint selection, and objective-specific changes followed label-free pretraining diagnostics (Appendix~\ref{app:config}).

\paragraph{Adaptation.}
Frozen probing trains a linear head for at most 50 epochs with learning rate $10^{-4}$ while keeping the encoder fixed. Full fine-tuning trains the encoder and a width-64 ReLU head for at most 100 epochs with learning rate $3\times10^{-4}$. Both use batch size 64, weight decay .05, cosine scheduling, gradient clipping at 1.0, and validation patience 10. Neural classification uses square-root-balanced class weights and label smoothing of .1. LOS trains with mean squared error but uses mean absolute error in days for evaluation and checkpoint selection. Classification checkpoint selection maximizes validation AUPRC. The two protocols differ in trainable scope, head capacity, learning rate, and epoch cap.

\paragraph{Converged and loss-matched probes.}
The capped probe is not trained to convergence, and its LOS training loss differs from the reported metric. We therefore refit all 180 core full-label frozen probes on the same embeddings, splits, and test rows: L2 logistic regression fitted to convergence, and for LOS a Huber probe approximating absolute error and a ridge probe. Validation selects regularization, and the protocol was fixed before any of its test metrics were computed (Appendix~\ref{app:converged}). Appendix~\ref{app:post-review} collects all analyses added during peer review. We recommend the converged probe for new objectives; its script is released with the diagnostics but is not yet a framework protocol.

\paragraph{Baselines.}
The supervised Transformer uses the same encoder family and full-fine-tuning trainer from random initialization. GRU-D provides a recurrent, missingness-aware comparator \citep{che2018grud}. XGBoost uses per-variable summary features, including extrema, first and last values, counts, and observation fractions \citep{chen2016xgboost}. TS2Vec is a temporal-contrastive extension, not a fourth core objective.

\paragraph{Seeds, metrics, and checkpoints.}
All reported core cells contain five runs with matched seeds. A seed jointly determines the patient split and stochastic training. We report mean and sample standard deviation across runs, without confirmatory significance claims (Section~\ref{sec:limitations}).

Every downstream SSL run starts from the encoder saved after the final pretraining epoch, not one selected on objective-specific validation loss. This gives all objectives the same 100-epoch data-pass rule but may disadvantage an objective whose useful representations peak earlier. Selecting encoders on downstream validation would make SSL checkpoint choice label-dependent and task-specific. The final epoch is a label-free rule that treats objectives alike. Downstream and baseline test metrics use validation-selected downstream checkpoints. Section~\ref{sec:secondary} tests this rule.
 
\section{Results}
\label{sec:results}

\subsection{Core Classification Results}

\tableref{tab:classification} reports all full-label classification cells. Under the capped linear probe, \MAE{} has the highest mean AUPRC in all nine cells. This ordering holds in all 90 matched-seed comparisons against JEPA and contrastive learning. Mean paired advantages over the stronger SSL alternative range from .026 to .084 AUPRC (Appendix Figure~\ref{fig:seed-differences}). Under fine-tuning, the SSL ranking is less uniform and paired differences often cross zero: \MAE{} leads seven cells, ties \JEPA{} on \Combined{} AKI (.626 each), and trails contrastive learning on \MIMIC{} 24-hour mortality. The supervised Transformer has the highest fine-tuned mean in all three AKI cells.

Probe optimization does not drive the frozen ranking (\tableref{tab:classification}; Appendix~\ref{app:converged}). The converged probe scores higher in 132 of 135 runs, by .017 AUPRC on average. Gains are largest for 24-hour mortality, where 82\% of capped probes stopped early on a low-prevalence validation metric, and .004--.012 per objective elsewhere, where 93--96\% reached the epoch cap. The objective order is unchanged in all nine cells by AUPRC and AUROC. \MAE{} beats both alternatives in all 90 matched-seed comparisons, with a mean paired advantage of .018--.081.

\begin{table*}[t]
\floatconts
  {tab:classification}
  {\caption{Core classification AUPRC, mean over five matched seeds. Capped probe: linear probe trained for at most 50 epochs. Converged probe: L2 logistic regression fitted to convergence on the same embeddings. Baselines are trained from scratch with full labels (Transf.: supervised Transformer). Bold marks the best mean per row within each probe block and across fine-tuning and baselines, ties at three decimals included. Appendix Tables~\ref{tab:classification-full} and~\ref{tab:converged-probe} give SDs.}}
  {\scriptsize
\setlength{\tabcolsep}{3.6pt}
\begin{tabular}{@{}llrrrrrrrrrrrr@{}}
\toprule
Data & Endpoint & \multicolumn{3}{c}{Probe, capped} & \multicolumn{3}{c}{Probe, converged} & \multicolumn{3}{c}{Fine-tune} & \multicolumn{3}{c}{Supervised baselines} \\
\cmidrule(lr){3-5} \cmidrule(lr){6-8} \cmidrule(lr){9-11} \cmidrule(lr){12-14}
 & & MAE & JEPA & Con. & MAE & JEPA & Con. & MAE & JEPA & Con. & Transf. & GRU-D & XGBoost \\
\midrule
Combined & 24h mortality & \textbf{0.284} & 0.188 & 0.205 & \textbf{0.309} & 0.208 & 0.239 & 0.409 & 0.391 & 0.399 & 0.389 & \textbf{0.422} & 0.374 \\
Combined & Hospital mortality & \textbf{0.470} & 0.357 & 0.392 & \textbf{0.470} & 0.363 & 0.395 & \textbf{0.531} & 0.521 & 0.514 & 0.518 & 0.527 & 0.525 \\
Combined & AKI & \textbf{0.447} & 0.327 & 0.396 & \textbf{0.451} & 0.332 & 0.402 & 0.626 & 0.626 & 0.614 & 0.652 & 0.630 & \textbf{0.668} \\
eICU-CRD & 24h mortality & \textbf{0.276} & 0.199 & 0.195 & \textbf{0.287} & 0.219 & 0.215 & 0.377 & 0.363 & 0.360 & 0.351 & \textbf{0.397} & 0.315 \\
eICU-CRD & Hospital mortality & \textbf{0.439} & 0.355 & 0.350 & \textbf{0.439} & 0.359 & 0.356 & \textbf{0.495} & 0.471 & 0.471 & 0.474 & 0.484 & 0.482 \\
eICU-CRD & AKI & \textbf{0.361} & 0.268 & 0.286 & \textbf{0.366} & 0.275 & 0.293 & 0.519 & 0.515 & 0.494 & 0.554 & 0.510 & \textbf{0.569} \\
MIMIC-IV & 24h mortality & \textbf{0.312} & 0.236 & 0.254 & \textbf{0.357} & 0.298 & 0.322 & 0.451 & 0.451 & 0.456 & 0.449 & \textbf{0.480} & 0.402 \\
MIMIC-IV & Hospital mortality & \textbf{0.510} & 0.415 & 0.458 & \textbf{0.520} & 0.440 & 0.471 & \textbf{0.579} & 0.564 & 0.568 & 0.565 & 0.577 & 0.577 \\
MIMIC-IV & AKI & \textbf{0.504} & 0.410 & 0.478 & \textbf{0.510} & 0.433 & 0.492 & 0.667 & 0.642 & 0.645 & 0.693 & 0.661 & \textbf{0.729} \\
\bottomrule
\end{tabular}
}
\end{table*}

\subsection{SSL Versus Supervised and Classical Baselines}

At full labels, baselines lead six of the twelve cells (\tableref{tab:classification} and the fine-tune rows of \tableref{tab:los}; Appendix \tableref{tab:baseline-full} gives the compact matrix). GRU-D has the highest 24-hour-mortality AUPRC and XGBoost the highest AKI AUPRC in all three regimes. Fine-tuned \MAE{} leads hospital mortality in all three, narrowly (.531, .495, and .579 against at most .527, .484, and .577 on \Combined{}, \EICU{}, and \MIMIC{}). The best SSL model has the lowest LOS error in all three regimes, by .03--.06 days. The baselines share cohorts, splits, and metrics but differ in input summaries and inductive biases.

\subsection{Length of Stay and Loss Matching}

The capped LOS probes trained with squared error but were selected and evaluated on absolute error. The target is right-skewed, with training means of 2.7--3.4 days and medians of 1.3--1.5 days. Absolute error therefore favors the training-median constant, while a squared-error model estimates the conditional mean. With an absolute-error loss, all three objectives beat the median constant in 5/5 seeds per regime, by 7--12\% (\tableref{tab:los}). Ridge probes lose to this constant on absolute error but beat a training-mean constant on squared error in 5/5 seeds. Objective differences are small: \MAE{} is lowest on \EICU{} and \Combined{}, and the three are within .023 days on \MIMIC{}, where \MAE{} is last in 5/5 seeds. The capped-probe ordering, in which contrastive learning was lowest and frozen \MAE{} was above the constant (Appendix \tableref{tab:los-reference}), does not survive loss matching. Its contrastive heads came from the first few epochs. Fine-tuning still trains with squared error and beats the constant in every regime, but was not rerun with an absolute-error loss.

\begin{table*}[t]
\floatconts
  {tab:los}
  {\caption{Remaining-LOS absolute error in days, mean (sample SD), $n=5$; lower is better. Probe rows are linear probes on frozen final-epoch embeddings, trained with an absolute-error-type Huber loss (abs.) or ridge squared error (sq.). Fine-tune rows are the reference squared-error runs beside full-label baselines (Transf.: supervised Transformer). Const.\ predicts each seed's training-label median. Bold marks the lowest learned-model mean per row; in sq.\ probe rows the constant is lower.}}
  {\scriptsize
\setlength{\tabcolsep}{2.5pt}
\begin{tabular}{@{}llrrrrrrr@{}}
\toprule
Data & Probe/protocol & MAE & JEPA & Con. & Transf. & GRU-D & XGBoost & Const. \\
\midrule
Combined & Probe, abs. & \textbf{2.193 (0.028)} & 2.215 (0.027) & 2.200 (0.031) & \multicolumn{1}{c}{--} & \multicolumn{1}{c}{--} & \multicolumn{1}{c}{--} & 2.403 (0.025) \\
Combined & Probe, sq. & \textbf{2.469 (0.025)} & 2.506 (0.024) & 2.486 (0.031) & \multicolumn{1}{c}{--} & \multicolumn{1}{c}{--} & \multicolumn{1}{c}{--} & 2.403 (0.025) \\
Combined & Fine-tune, sq. & \textbf{2.118 (0.029)} & 2.135 (0.013) & 2.152 (0.024) & 2.190 (0.031) & 2.174 (0.031) & 2.340 (0.020) & 2.403 (0.025) \\
eICU-CRD & Probe, abs. & \textbf{1.992 (0.010)} & 2.024 (0.011) & 2.018 (0.008) & \multicolumn{1}{c}{--} & \multicolumn{1}{c}{--} & \multicolumn{1}{c}{--} & 2.176 (0.006) \\
eICU-CRD & Probe, sq. & \textbf{2.228 (0.011)} & 2.283 (0.015) & 2.272 (0.006) & \multicolumn{1}{c}{--} & \multicolumn{1}{c}{--} & \multicolumn{1}{c}{--} & 2.176 (0.006) \\
eICU-CRD & Fine-tune, sq. & \textbf{1.964 (0.017)} & 1.977 (0.034) & 2.008 (0.031) & 1.997 (0.030) & 2.003 (0.018) & 2.163 (0.015) & 2.176 (0.006) \\
MIMIC-IV & Probe, abs. & 2.439 (0.062) & 2.420 (0.059) & \textbf{2.416 (0.062)} & \multicolumn{1}{c}{--} & \multicolumn{1}{c}{--} & \multicolumn{1}{c}{--} & 2.733 (0.074) \\
MIMIC-IV & Probe, sq. & 2.774 (0.046) & \textbf{2.750 (0.048)} & 2.757 (0.051) & \multicolumn{1}{c}{--} & \multicolumn{1}{c}{--} & \multicolumn{1}{c}{--} & 2.733 (0.074) \\
MIMIC-IV & Fine-tune, sq. & 2.368 (0.051) & 2.371 (0.069) & \textbf{2.359 (0.069)} & 2.397 (0.066) & 2.394 (0.046) & 2.613 (0.038) & 2.733 (0.074) \\
\bottomrule
\end{tabular}
}
\end{table*}

\subsection{Label Efficiency}
\label{sec:label-efficiency}

Does pretraining help when labels are scarce? We reduce only the training labels, to 1--50\% for hospital mortality, 5--50\% for 24-hour mortality, and 10\% for AKI and LOS, giving 33 reduced-label cells per protocol. Among the three SSL implementations, \MAE{} leads 30/33 fine-tuning and 28/33 capped-probe means (28/30 classification cells under each protocol), and 28/33 fine-tuning means with the supervised Transformer included. For hospital mortality, fine-tuned \MAE{} exceeds the supervised Transformer at every fraction in every regime, by .033 (paired SD .012) at 1\% labels on \Combined{}. No single fraction marks where pretraining stops paying off: for \MIMIC{} 24-hour mortality, differences above 5\% labels change sign, and at 5\% the unstratified subsets have 31--43 training positives per seed. Reduced-label GRU-D and XGBoost runs are not compared. Appendix Figure~\ref{fig:label-efficiency} and \tableref{tab:label-fraction-delta} give all fractions.

\subsection{Secondary Diagnostics}
\label{sec:secondary}

\paragraph{Sources and databases.}
Scored separately on the \MIMIC{} and \EICU{} portions of the \Combined{} test split, converged \MAE{} probes lead both alternatives on both portions for all classification tasks in 5/5 seeds. Fine-tuned \MAE{} leads five of six portion--task means among SSL objectives, and supervised training leads AKI on both portions. Pooling hides a difference in absolute level, with higher AUROC on the \MIMIC{} portion, not a reversal (Appendix~\ref{app:per-source}). With source pretraining and target-label fine-tuning, MAE leads or ties all four \MIMIC{}-to-\EICU{} tasks and leads three reverse tasks, and stays within .004--.010 AUPRC of target pretraining for mortality (Appendix~\ref{app:transfer}). This is adaptation with target labels, not out-of-domain validation (Section~\ref{sec:limitations}).

\paragraph{Compute.}
The 45 core pretraining runs took an estimated 55--63 L4-hours. \JEPA{} costs the most on every GPU measure, with 2.2 times the forward FLOPs of \MAE{} and 1.5 times its idle-GPU step time, yet ranks last in 7 of 9 frozen classification cells. \MAE{} uses the fewest FLOPs, while a contrastive step takes 18\% less time on an idle GPU. Cost alone therefore does not explain the ranking, although we ran no compute-matched ablation (Appendix~\ref{app:compute}).

\paragraph{Checkpoint selection.}
Selecting \MIMIC{} encoders on SSL validation loss instead of the final epoch worsens \JEPA{} capped probes on hospital mortality and LOS by .041 AUPRC and .26 days, changes \MAE{} and contrastive learning by at most .001, and leaves both leaders unchanged (Appendix~\ref{app:diagnostics}).

\paragraph{Missingness.}
A linear control on \MIMIC{} hospital mortality reaches .301 (.025) AUPRC from masks alone and .404 (.013) from values plus masks, evidence of input signal, not an optimized baseline (Appendix~\ref{app:raw}).

\paragraph{Other diagnostics.}
Medium and large \MAE{} encoders beat matched supervised models on \MIMIC{} 24-hour mortality at three label fractions, and a random encoder trails SSL probes in 118/120 seed-level comparisons. Appendices~\ref{app:capacity-subgroups}--\ref{app:random} add the TS2Vec-style extension and subgroup audits.
 
\section{Discussion}
\label{sec:discussion}

\paragraph{What can be reused.}
Researchers can audit run aggregates or add an objective while retaining clinical inputs, encoder transfer, and evaluation. A new objective then faces two references: the three SSL implementations and the supervised, recurrent, and tabular baselines, which lead half of the full-label cells.

\paragraph{Why reconstruction may fit the interface.}
The tensor exposes a value and observation mask at each hour. MAE reconstructs observed cells, JEPA predicts an evolving target embedding, and core contrastive learning compares pooled stay vectors. An interface preference for reconstruction is one hypothesis for the classification ordering. The controls do not identify the learned information or establish that continuous values are better than quantization.

\paragraph{Interpreting missingness.}
Measurement patterns carry outcome signal: masks alone give a linear model .301 AUPRC for \MIMIC{} hospital mortality at 11.7\% prevalence. The signal is ambiguous, because an unmeasured laboratory value may reflect clinical judgment, workflow, availability, or data loss, so a model can learn site-specific measurement policy, not only physiology. Preserving the mask avoids treating missing cells as observed zeros but does not make missingness causal or transportable. The frozen classification ranking held on both source portions of \Combined{}, which does not show transfer to a new hospital.

\paragraph{Evaluation practice.}
Both probes and fine-tuning remain necessary because they test different representation properties, and baselines show whether an SSL ordering matters for task performance. Frozen probes should be trained to convergence, and regression probes should use a loss matched to the reported metric. Checkpoint and cost policies must be explicit, because equal epochs do not imply equal compute and checkpoint selection affected JEPA.
 
\section{Limitations}
\label{sec:limitations}

The benchmark holds clinical inputs and downstream evaluation fixed while retaining the components needed by each objective. Objective-specific heads, target networks, encoder passes, and trainable parameter counts therefore differ, so results describe these implementations, not objective families. Configurations reflect unequal development histories and a small targeted sensitivity analysis, not matched compute or tuning budgets or objective-specific optima, and the benchmark was not preregistered.

\paragraph{No out-of-domain validation.}
No model is evaluated on a database or hospital without training on its labels. Cross-database analyses fine-tune on target labels, and per-source results stay within the pooled regime. The results therefore do not show that rankings or predictors hold in a new hospital, for instance one with different ordering practices. Such a test needs source normalization applied to target data, which the benchmark does not yet provide. \Combined{} is a union, not an external cohort, and both databases are from the US, retrospective, and deidentified.

\paragraph{Representation and architecture.}
Rankings may change with observation-triplet, quantized, event-token, or patch representations, which also change the masking unit. Waveforms, notes, images, and full medication histories are out of scope. All objectives use one small Transformer. SSL on decay-based recurrent encoders such as GRU-D is the first architecture extension, since masked hours must then be skipped or fed as missing. Shared masked views may underrepresent contrastive methods with clinically defined positive pairs.

\paragraph{Statistics and labels.}
All comparisons are descriptive. Five matched seeds are few and reuse overlapping source populations. They describe split and optimization variation but do not establish population uncertainty, statistical equivalence, calibration quality, subgroup fairness, clinical utility, or deployment readiness. Reduced-label subsets are not class-stratified, and reduced-label GRU-D and XGBoost runs are not compared. The AKI label uses creatinine only. Fine-tuned LOS models train with squared error. Final-epoch checkpoints may miss better earlier states. The random-encoder, raw/mask, and post-review diagnostics are post hoc.
 
\section{Ethics, Access, and Reproducibility}
\label{sec:ethics}

The study analyzes retrospective, deidentified ICU records under the credentialing and data-use terms of PhysioNet. It involves no prospective intervention, patient contact, or automated clinical decision. Benchmark results come from the unchanged archived run aggregates; diagnostics are stored separately.

We reran a \MIMIC{} slice with the maintained framework, starting from the processed tensors (Appendix~\ref{app:reproduction}). From archived reference encoders, 12 frozen probes reproduce every reference test metric to numerical precision, and 12 fine-tuning runs lie within one across-seed SD. The full pipeline, with 6 pretraining and 48 downstream runs over two seeds, differs from the reference by .008 mean absolute AUPRC for probes and .010 for fine-tuning, all within two SDs, and keeps the objective order in 13 of 16 cells. LOS fine-tuning error was higher in all nine comparisons, by .03--.04 days on average, for reasons we have not identified, and other databases and seeds were not rerun.

\section{Conclusion}
\label{sec:conclusion}

\SLICES{} provides a fixed sparse-ICU benchmark specification, interchangeable SSL objectives, shared downstream evaluation, and auditable reference results. At full labels, baselines lead half of the cells: GRU-D leads 24-hour mortality and XGBoost leads AKI in every regime, while fine-tuned SSL leads hospital mortality narrowly and LOS. Among the SSL objectives, MAE has the highest mean AUPRC in all nine frozen cells under both probes and in 28 of 30 reduced-label fine-tuning classification cells, while LOS differences are small once the probe loss matches the metric. Generalization to unseen databases remains untested. Future work will add an out-of-domain protocol, make converged and loss-matched probes framework protocols, and test the objectives on recurrent encoders such as GRU-D, on other input representations, and under matched compute.

\acks{We thank the anonymous reviewers and the area chair for comments that led to the converged-probe, loss-matched LOS, per-source, compute, and reproduction analyses. LAC is funded by the National Institutes of Health through DS-I Africa U54 TW012043-01 and Bridge2AI OT2OD032701, the National Science Foundation through ITEST \#2148451, a grant of the Boston-Korea Innovative Research Project (RS-2024-00403047), and a grant of the Korea Health Technology R\&D Project (RS-2024-00439677) through the Korea Health Industry Development Institute (KHIDI), funded by the Ministry of Health \& Welfare, Republic of Korea.}

\bibliography{references}

\begin{thebibliography}{26}
\providecommand{\natexlab}[1]{#1}
\providecommand{\url}[1]{\texttt{#1}}
\expandafter\ifx\csname urlstyle\endcsname\relax
  \providecommand{\doi}[1]{doi: #1}\else
  \providecommand{\doi}{doi: \begingroup \urlstyle{rm}\Url}\fi

\bibitem[Assran et~al.(2023)Assran, Duval, Misra, Bojanowski, Vincent, Rabbat,
  LeCun, and Ballas]{assran2023ijepa}
Mahmoud Assran, Quentin Duval, Ishan Misra, Piotr Bojanowski, Pascal Vincent,
  Michael Rabbat, Yann LeCun, and Nicolas Ballas.
\newblock Self-supervised learning from images with a joint-embedding
  predictive architecture.
\newblock In \emph{IEEE/CVF Conference on Computer Vision and Pattern
  Recognition}, pages 15619--15629, 2023.
\newblock \doi{10.1109/CVPR52729.2023.01499}.

\bibitem[Bennett et~al.(2023)Bennett, Plecko, Ukor, Meinshausen, and
  B{\"u}hlmann]{bennett2023ricu}
Nicolas Bennett, Drago Plecko, Ida-Fong Ukor, Nicolai Meinshausen, and Peter
  B{\"u}hlmann.
\newblock ricu: {R}'s interface to intensive care data.
\newblock \emph{GigaScience}, 12:\penalty0 giad041, 2023.
\newblock \doi{10.1093/gigascience/giad041}.

\bibitem[Che et~al.(2018)Che, Purushotham, Cho, Sontag, and Liu]{che2018grud}
Zhengping Che, Sanjay Purushotham, Kyunghyun Cho, David Sontag, and Yan Liu.
\newblock Recurrent neural networks for multivariate time series with missing
  values.
\newblock \emph{Scientific Reports}, 8:\penalty0 6085, 2018.
\newblock \doi{10.1038/s41598-018-24271-9}.

\bibitem[Chen and Guestrin(2016)]{chen2016xgboost}
Tianqi Chen and Carlos Guestrin.
\newblock {XGBoost}: A scalable tree boosting system.
\newblock In \emph{Proceedings of the 22nd ACM SIGKDD International Conference
  on Knowledge Discovery and Data Mining}, pages 785--794, 2016.
\newblock \doi{10.1145/2939672.2939785}.

\bibitem[Chen et~al.(2020)Chen, Kornblith, Norouzi, and Hinton]{chen2020simclr}
Ting Chen, Simon Kornblith, Mohammad Norouzi, and Geoffrey Hinton.
\newblock A simple framework for contrastive learning of visual
  representations.
\newblock In \emph{Proceedings of the 37th International Conference on Machine
  Learning}, volume 119 of \emph{Proceedings of Machine Learning Research},
  pages 1597--1607, 2020.
\newblock URL \url{https://proceedings.mlr.press/v119/chen20j.html}.

\bibitem[Ennadir et~al.(2024)Ennadir, Golkar, and Sarra]{ennadir2024tsjepa}
Sofiane Ennadir, Siavash Golkar, and Leopoldo Sarra.
\newblock Joint embeddings go temporal.
\newblock In \emph{NeurIPS Workshop on Time Series in the Age of Large Models},
  2024.
\newblock URL \url{https://openreview.net/forum?id=FIdbozebmy}.

\bibitem[Guo et~al.(2026)Guo, Arciniegas, Yan, Fries, Tomlinson, and
  Sung]{guo2026ehrfoundationreview}
Lin~Lawrence Guo, Santiago~Eduardo Arciniegas, Adam~Paul Yan, Jason Fries,
  George~A. Tomlinson, and Lillian Sung.
\newblock Systematic review of foundation models for structured electronic
  health records.
\newblock \emph{Journal of the American Medical Informatics Association},
  33\penalty0 (6):\penalty0 1190--1198, 2026.
\newblock \doi{10.1093/jamia/ocag033}.

\bibitem[Harutyunyan et~al.(2019)Harutyunyan, Khachatrian, Kale, Ver~Steeg, and
  Galstyan]{harutyunyan2019multitask}
Hrayr Harutyunyan, Hrant Khachatrian, David~C. Kale, Greg Ver~Steeg, and Aram
  Galstyan.
\newblock Multitask learning and benchmarking with clinical time series data.
\newblock \emph{Scientific Data}, 6:\penalty0 96, 2019.
\newblock \doi{10.1038/s41597-019-0103-9}.

\bibitem[He et~al.(2022)He, Chen, Xie, Li, Doll{\'a}r, and Girshick]{he2022mae}
Kaiming He, Xinlei Chen, Saining Xie, Yanghao Li, Piotr Doll{\'a}r, and Ross
  Girshick.
\newblock Masked autoencoders are scalable vision learners.
\newblock In \emph{IEEE/CVF Conference on Computer Vision and Pattern
  Recognition}, pages 16000--16009, 2022.
\newblock \doi{10.1109/CVPR52688.2022.01553}.

\bibitem[Johnson et~al.(2024)Johnson, Bulgarelli, Pollard, Gow, Moody, Horng,
  Celi, and Mark]{physionet2024mimiciv31}
Alistair Johnson, Lucas Bulgarelli, Tom Pollard, Brian Gow, Benjamin Moody,
  Steven Horng, Leo~Anthony Celi, and Roger Mark.
\newblock {MIMIC-IV} (version 3.1), 2024.
\newblock URL \url{https://doi.org/10.13026/kpb9-mt58}.

\bibitem[Johnson et~al.(2023)Johnson, Bulgarelli, Shen, Gayles, Shammout,
  Horng, Pollard, Hao, Moody, Gow, Lehman, Celi, and Mark]{johnson2023mimiciv}
Alistair E.~W. Johnson, Lucas Bulgarelli, Lu~Shen, Alvin Gayles, Ayad Shammout,
  Steven Horng, Tom~J. Pollard, Sicheng Hao, Benjamin Moody, Brian Gow,
  Li-wei~H. Lehman, Leo~Anthony Celi, and Roger~G. Mark.
\newblock {MIMIC-IV}, a freely accessible electronic health record dataset.
\newblock \emph{Scientific Data}, 10\penalty0 (1), 2023.
\newblock \doi{10.1038/s41597-022-01899-x}.

\bibitem[{Kidney Disease: Improving Global Outcomes Acute Kidney Injury Work
  Group}(2012)]{kdigo2012aki}
{Kidney Disease: Improving Global Outcomes Acute Kidney Injury Work Group}.
\newblock {KDIGO} clinical practice guideline for acute kidney injury.
\newblock \emph{Kidney International Supplements}, 2\penalty0 (1):\penalty0
  1--138, 2012.
\newblock URL
  \url{https://kdigo.org/wp-content/uploads/2016/10/KDIGO-2012-AKI-Guideline-English.pdf}.

\bibitem[Labach et~al.(2023)Labach, Pokhrel, Huang, Zuberi, Yi, Volkovs,
  Poutanen, and Krishnan]{labach2023duett}
Alex Labach, Aslesha Pokhrel, Xiao~Shi Huang, Saba Zuberi, Seung~Eun Yi,
  Maksims Volkovs, Tomi Poutanen, and Rahul~G. Krishnan.
\newblock {DuETT}: Dual event time transformer for electronic health records.
\newblock In \emph{Proceedings of the Machine Learning for Healthcare
  Conference}, volume 219 of \emph{Proceedings of Machine Learning Research},
  pages 403--422, 2023.
\newblock URL \url{https://proceedings.mlr.press/v219/labach23a.html}.

\bibitem[Oufattole et~al.(2024)Oufattole, Jeong, McDermott, Balagopalan,
  Jangeesingh, Ghassemi, and Stultz]{oufattole2024ebcl}
Nassim Oufattole, Hyewon Jeong, Matthew B.~A. McDermott, Aparna Balagopalan,
  Bryan Jangeesingh, Marzyeh Ghassemi, and Collin Stultz.
\newblock Event-based contrastive learning for medical time series.
\newblock In \emph{Proceedings of the Machine Learning for Healthcare
  Conference}, volume 252 of \emph{Proceedings of Machine Learning Research},
  2024.
\newblock URL \url{https://proceedings.mlr.press/v252/oufattole24a.html}.

\bibitem[Pollard et~al.(2019)Pollard, Johnson, Raffa, Celi, Badawi, and
  Mark]{physionet2019eicu20}
Tom Pollard, Alistair Johnson, Jesse Raffa, Leo~Anthony Celi, Omar Badawi, and
  Roger Mark.
\newblock {eICU} collaborative research database (version 2.0), 2019.
\newblock URL \url{https://doi.org/10.13026/C2WM1R}.

\bibitem[Pollard et~al.(2018)Pollard, Johnson, Raffa, Celi, Mark, and
  Badawi]{pollard2018eicu}
Tom~J. Pollard, Alistair E.~W. Johnson, Jesse~D. Raffa, Leo~Anthony Celi,
  Roger~G. Mark, and Omar Badawi.
\newblock The {eICU} collaborative research database, a freely available
  multi-center database for critical care research.
\newblock \emph{Scientific Data}, 5:\penalty0 180178, 2018.
\newblock \doi{10.1038/sdata.2018.178}.

\bibitem[Raghu et~al.(2023)Raghu, Chandak, Alam, Guttag, and
  Stultz]{raghu2023smdssl}
Aniruddh Raghu, Payal Chandak, Ridwan Alam, John Guttag, and Collin Stultz.
\newblock Sequential multi-dimensional self-supervised learning for clinical
  time series.
\newblock In \emph{Proceedings of the 40th International Conference on Machine
  Learning}, volume 202 of \emph{Proceedings of Machine Learning Research},
  pages 28531--28548, 2023.
\newblock URL \url{https://proceedings.mlr.press/v202/raghu23a.html}.

\bibitem[Saito and Rehmsmeier(2015)]{saito2015precisionrecall}
Takaya Saito and Marc Rehmsmeier.
\newblock The precision-recall plot is more informative than the {ROC} plot
  when evaluating binary classifiers on imbalanced datasets.
\newblock \emph{PLOS ONE}, 10\penalty0 (3):\penalty0 e0118432, 2015.
\newblock \doi{10.1371/journal.pone.0118432}.

\bibitem[Tan et~al.(2023)Tan, Getzen, Hutch, Strasser,
  et~al.]{tan2023informativemissingness}
Amelia L.~M. Tan, Emily~J. Getzen, Meghan~R. Hutch, Zachary~H. Strasser, et~al.
\newblock Informative missingness: What can we learn from patterns in missing
  laboratory data in the electronic health record?
\newblock \emph{Journal of Biomedical Informatics}, 139:\penalty0 104306, 2023.
\newblock \doi{10.1016/j.jbi.2023.104306}.

\bibitem[Tipirneni and Reddy(2022)]{tipirneni2022strats}
Sindhu Tipirneni and Chandan~K. Reddy.
\newblock Self-supervised transformer for sparse and irregularly sampled
  multivariate clinical time-series.
\newblock \emph{ACM Transactions on Knowledge Discovery from Data}, 16\penalty0
  (6):\penalty0 105, 2022.
\newblock \doi{10.1145/3516367}.

\bibitem[van~de Water et~al.(2024)van~de Water, Schmidt, Elbers, Thoral,
  Arnrich, and Rockenschaub]{van_de_water2024yaib}
Robin van~de Water, Hendrik Schmidt, Paul Elbers, Patrick Thoral, Bert Arnrich,
  and Patrick Rockenschaub.
\newblock Yet another {ICU} benchmark: A flexible multi-center framework for
  clinical {ML}.
\newblock In \emph{International Conference on Learning Representations}, 2024.

\bibitem[Xiang et~al.(2026)Xiang, Restrepo, Jeong, Jia, and
  Celi]{xiang2026aidmae}
Xiao Xiang, David Restrepo, Hyewon Jeong, Yugang Jia, and Leo~Anthony Celi.
\newblock Learning representations from incomplete {EHR} data with dual-masked
  autoencoding.
\newblock \emph{arXiv preprint arXiv:2602.15159}, 2026.
\newblock \doi{10.48550/arXiv.2602.15159}.

\bibitem[Yang et~al.(2026)Yang, Arora, Zhang, Abed, Kim, Choudhary,
  Hassanuzzaman, Zhu, Ali, Yang, Gent, Moas, and
  Kamaleswaran]{yang2026clinjepa}
Yixuan Yang, Mehak Arora, Ryan Zhang, Baraa Abed, Junseob Kim, Tilendra
  Choudhary, Md~Hassanuzzaman, Kevin Zhu, Ayman Ali, Chengkun Yang,
  Alasdair~Edward Gent, Victor Moas, and Rishikesan Kamaleswaran.
\newblock {Clin-JEPA}: A multi-phase co-training framework for joint-embedding
  predictive pretraining on {EHR} patient trajectories.
\newblock \emph{arXiv preprint arXiv:2605.10840}, 2026.
\newblock \doi{10.48550/arXiv.2605.10840}.

\bibitem[Y{\`e}che et~al.(2021)Y{\`e}che, Dresdner, Locatello, H{\"u}ser, and
  R{\"a}tsch]{yeche2021ncl}
Hugo Y{\`e}che, Gideon Dresdner, Francesco Locatello, Matthias H{\"u}ser, and
  Gunnar R{\"a}tsch.
\newblock Neighborhood contrastive learning applied to online patient
  monitoring.
\newblock In \emph{Proceedings of the 38th International Conference on Machine
  Learning}, volume 139 of \emph{Proceedings of Machine Learning Research},
  pages 11964--11974, 2021.
\newblock URL \url{https://proceedings.mlr.press/v139/yeche21a.html}.

\bibitem[Yu et~al.(2024)Yu, Chu, Jin, Wang, and Zhao]{yu2024smart}
Zhihao Yu, Xu~Chu, Yujie Jin, Yasha Wang, and Junfeng Zhao.
\newblock {SMART}: Towards pre-trained missing-aware model for patient health
  status prediction.
\newblock In \emph{Advances in Neural Information Processing Systems},
  volume~37, pages 63986--64009, 2024.
\newblock \doi{10.52202/079017-2043}.

\bibitem[Yue et~al.(2022)Yue, Wang, Duan, Yang, Huang, Tong, and
  Xu]{yue2022ts2vec}
Zhihan Yue, Yujing Wang, Juanyong Duan, Tianmeng Yang, Congrui Huang, Yunhai
  Tong, and Bixiong Xu.
\newblock {TS2Vec}: Towards universal representation of time series.
\newblock In \emph{Proceedings of the AAAI Conference on Artificial
  Intelligence}, volume~36, pages 8980--8987, 2022.
\newblock \doi{10.1609/aaai.v36i8.20881}.

\end{thebibliography}

\clearpage
\appendix
\section{Additional Benchmark Details}
\label{app:benchmark}

\subsection{Benchmark and Data Flow}

Figure~\ref{fig:data-pipeline} details the preprocessing path introduced in Figure~\ref{fig:thesis-overview}. Combined pools the two source databases. It is not an external cohort. The varying unit includes objective-specific masks, auxiliary modules, and compute.

\begin{figure*}[p]
\floatconts{fig:data-pipeline}
{\caption{Data pipeline. RICU supplies concept-specific harmonization and hourly aggregation. SLICES applies cohort and feature filters and forms values, masks, and labels. Patient assignment precedes task-specific null-label filtering. Normalization uses observed training cells, followed by zero filling with the observation mask retained. Event rows are illustrative.}}
{\includegraphics[width=\textwidth]{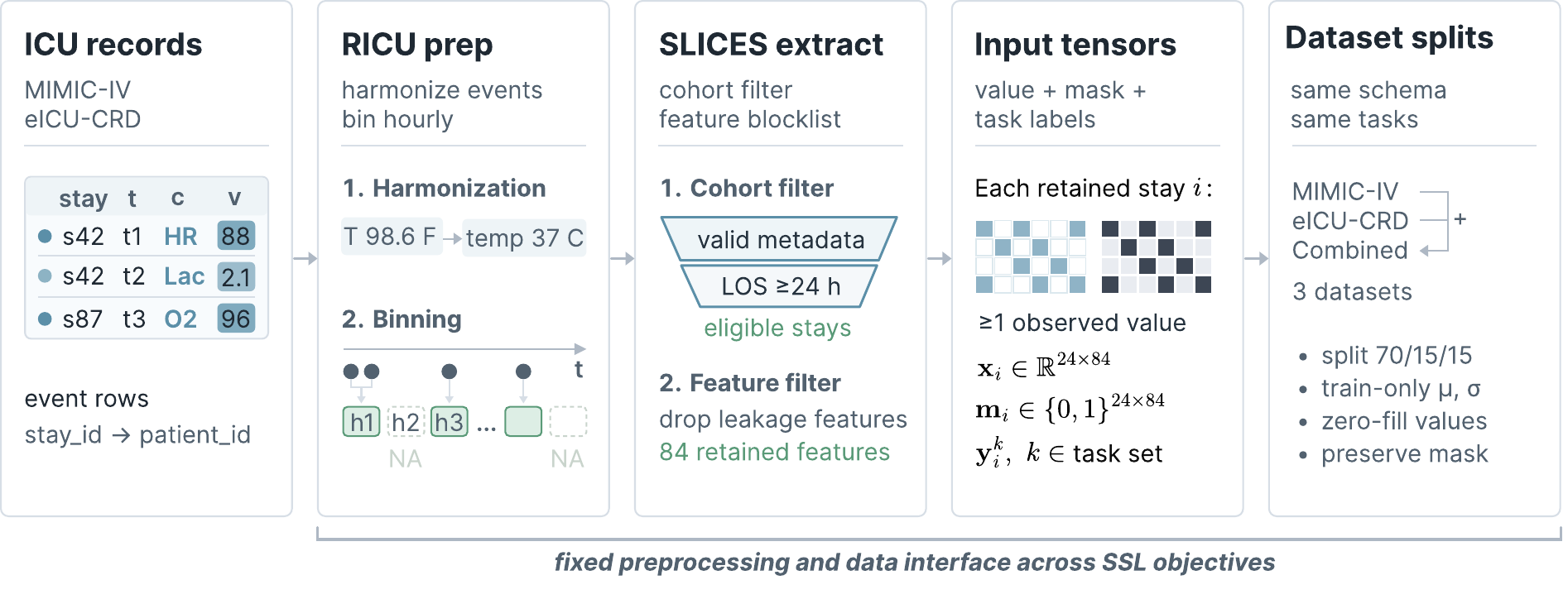}}
\end{figure*}

\begin{figure*}[p]
\floatconts{fig:encoder-tokenization}
{\caption{Shared hourly tokenization and encoder. Values and masks form 168 inputs per hour, projected to 64 dimensions. Empty hours are excluded from attention and losses. SSL uses timestep outputs, with objective-specific pooling for contrastive learning. Downstream heads mean-pool valid hours.}}
{\includegraphics[width=\textwidth]{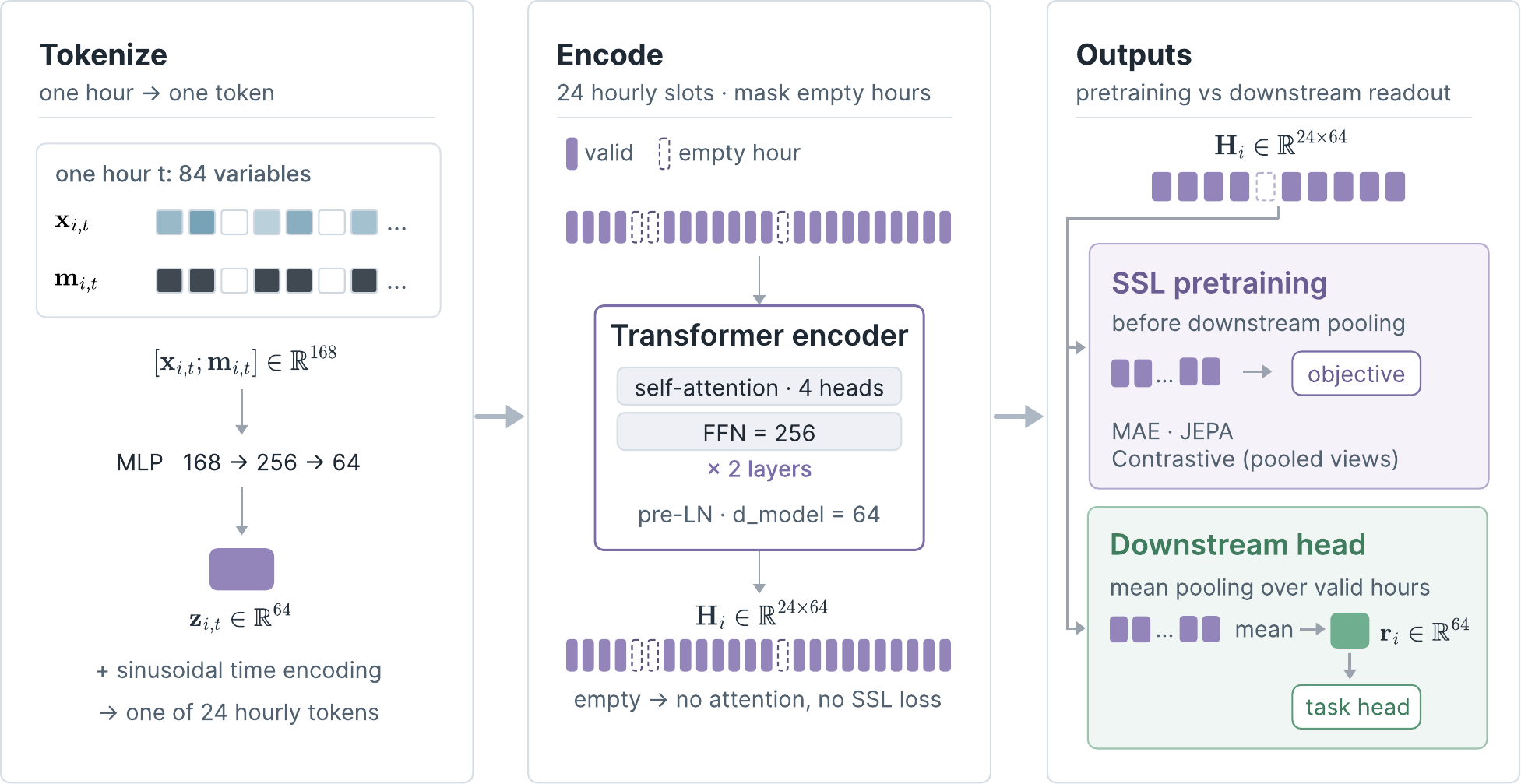}}
\end{figure*}

\subsection{Downstream Protocols}

Figure~\ref{fig:protocol-ab} separates frozen probing from full fine-tuning. Both use the same final pretraining encoder but differ in trainable scope, head capacity, learning rate, and epoch cap. These are two evaluation protocols, not a freeze-only ablation. The separate checkpoint diagnostic changes the SSL encoder selection policy only within its stated scope.

\begin{figure*}[p]
\floatconts{fig:protocol-ab}
{\caption{Reference downstream protocols. The displayed 50 and 100 epochs are maximum budgets. Validation patience is ten epochs. Protocol A (frozen probing) trains a linear head. Protocol B (full fine-tuning) adapts the encoder and a width-64 MLP head. Validation selects each downstream checkpoint before one test evaluation. Both start from final-epoch SSL weights.}}
{\includegraphics[width=\textwidth]{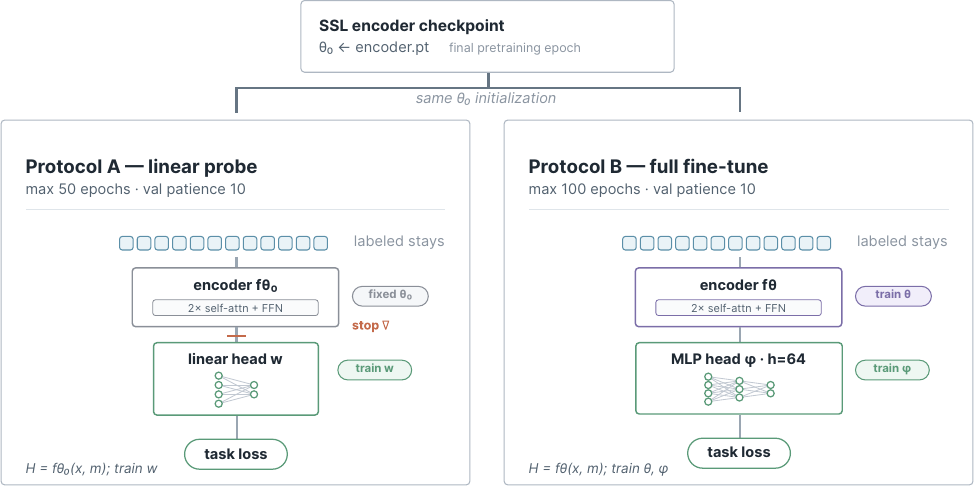}}
\end{figure*}

\subsection{Task Definitions and Test Support}

Tables~\ref{tab:task-definitions} and~\ref{tab:test-support} give the target rules and exported test support. Twenty-four-hour mortality and remaining LOS are fixed-window analogues of the decompensation and length-of-stay tasks of \citet{harutyunyan2019multitask}, and mortality, AKI, and LOS appear in YAIB \citep{van_de_water2024yaib}. The four endpoints span 1.6\%--20.3\% prevalence, two horizons, a laboratory-defined label with structured missing labels, and a continuous target. The SSL ordering is uniform for frozen classification but not for fine-tuning or against supervised and classical baselines, so these differences matter for the comparison.

\begin{table*}[htbp]
\floatconts
  {tab:task-definitions}
  {\caption{Downstream task contract. All tasks use hours 0--24 as model input.}}
  {\scriptsize
  \begin{tabular}{>{\raggedright\arraybackslash}p{.12\textwidth}>{\raggedright\arraybackslash}p{.12\textwidth}>{\raggedright\arraybackslash}p{.22\textwidth}>{\raggedright\arraybackslash}p{.25\textwidth}l}
  \toprule
  Task & Horizon after input & Target & Null-label rule & Primary metric \\
  \midrule
  24h mortality & hours 24--48 & death in forward day & death in hours 0--24 & AUPRC \\
  Hosp. mortality & after hour 24 & death before discharge & death in hours 0--24 & AUPRC \\
  AKI & hours 24--48 & creatinine-defined KDIGO~stage~$\geq$1 & missing baseline/forward evidence & AUPRC \\
  Remaining LOS & after hour 24 & ICU LOS minus one day, clipped at 0 & none & Absolute error (days) \\
  \bottomrule
  \end{tabular}}
\end{table*}

\begin{table*}[htbp]
\floatconts
  {tab:test-support}
  {\caption{Mean test support across the five patient splits. Positive counts and prevalence apply to classification tasks.}}
  {\scriptsize
\begin{tabular}{llrrr}
\toprule
Data & Task & Test samples & Positive & Prevalence \\
\midrule
Combined & 24h mortality & 30,863.4 & 492.4 & 0.0160 \\
Combined & Hospital mortality & 30,869.0 & 3,135.8 & 0.1016 \\
Combined & AKI & 21,724.0 & 3,565.2 & 0.1641 \\
Combined & Remaining LOS & 31,092.2 & -- & -- \\
eICU-CRD & 24h mortality & 19,788.6 & 315.2 & 0.0159 \\
eICU-CRD & Hospital mortality & 19,788.6 & 1,823.8 & 0.0922 \\
eICU-CRD & AKI & 13,297.4 & 1,850.0 & 0.1391 \\
eICU-CRD & Remaining LOS & 19,966.4 & -- & -- \\
MIMIC-IV & 24h mortality & 11,150.0 & 185.4 & 0.0166 \\
MIMIC-IV & Hospital mortality & 11,157.0 & 1,310.0 & 0.1174 \\
MIMIC-IV & AKI & 8,423.2 & 1,710.6 & 0.2031 \\
MIMIC-IV & Remaining LOS & 11,194.8 & -- & -- \\
\bottomrule
\end{tabular}
}
\end{table*}

\subsection{Objective and Downstream Configuration}
\label{app:config}

\begin{table}[!htbp]
\floatconts
  {tab:pretraining}
  {\caption{Core pretraining contract. Ratio is the configured timestep-mask ratio. All rows use AdamW, batch size 256, 100 epochs, and the same transferred encoder.}}
  {\scriptsize
  \begin{tabular}{llll}
  \toprule
  Objective & Masking & Target & Ratio \\
  \midrule
  \MAE{} & random hours & observed values & .5 \\
  \JEPA{} & $\leq3$ blocks & target embeddings & .5 \\
  Contrastive & two views & view agreement & .5 \\
  \bottomrule
  \end{tabular}}
\end{table}

\begin{table*}[htbp]
\floatconts
  {tab:objective-details}
  {\caption{Objective-specific modules and transferred capacity. Shared settings are listed in Section~\ref{sec:setup}.}}
  {\scriptsize
  \begin{tabular}{llllrr}
  \toprule
  Objective & Geometry & Learning target & Auxiliary module & Trainable & Transferred \\
  \midrule
  \MAE{} & random masked hours & observed normalized cells & Transformer decoder & 269,460 & 159,808 \\
  \JEPA{} & up to three masked blocks & EMA target embeddings & predictor + target encoder & 189,440 & 159,808 \\
  Contrastive & complementary views & instance agreement & projection MLP & 193,408 & 159,808 \\
  \bottomrule
  \end{tabular}}
\end{table*}

\paragraph{Development and test-set use.}
Every downstream run reports test metrics automatically, and the pipeline was run end to end a few times before the final rerun. Those earlier passes exposed defects that were fixed: an AKI label that overlapped the input window (detected through test AUROC above .97 for every method), mislabeled boundary cases in the mortality labels, and input features derived from future treatment durations. The test metrics we saw therefore came largely from earlier benchmark versions with different labels or inputs. Pretraining does not read the test partition, checkpoints are selected on validation metrics, and each run scores the test partition once after selection. The recorded objective-specific changes responded to label-free pretraining diagnostics: a rising JEPA loss with low representation spread, and a contrastive pretext task that was solved almost perfectly. Settings shaped by downstream comparisons are shared by all three objectives: input representation, encoder, one pretraining learning rate, one mask ratio, and one downstream recipe that the supervised Transformer also uses. Across the tested learning rates and mask ratios, mean test AUPRC moves by at most .018 within a sweep and .020 within an objective (\tableref{tab:hp}).

The \JEPA{} target encoder adds a non-trainable 159,808-parameter copy. Its EMA coefficient increases linearly from .999 to 1.0. The predictor has width 32, two layers, four heads, and feed-forward width 128. The contrastive projection has hidden width 256 and output width 64. All auxiliary modules are discarded before downstream evaluation.

\subsection{Retained Features and Preprocessing}

The ordered 84 RICU concept identifiers below define the model schema, not a claim that each is observed in both sources. Source-specific missingness remains explicit. Extraction uses RICU 0.6.3 concept definitions and unit conversions, requested at one-hour intervals. Concept-specific aggregation is inherited from RICU. Python tensorization neither quantizes values nor forward-fills across hours. It uses the observed training-partition mean and sample SD per feature. SD is floored at $10^{-6}$. Features absent in training retain a zero mean and unit SD. Missing values become zero after normalization. Low-label subsets retain the full training-partition normalizer. The code repository includes extraction scripts and the complete input blocklist, including direct LOS, future vasopressor-duration, and derived score features.

\begin{quote}\footnotesize
\texttt{abx}, \texttt{adh\_rate}, \texttt{alb}, \texttt{alp}, \texttt{alt}, \texttt{ast}, \texttt{avpu}, \texttt{basos}, \texttt{be}, \texttt{bicar}, \texttt{bili}, \texttt{bili\_dir}, \texttt{bnd}, \texttt{bun}, \texttt{ca}, \texttt{cai}, \texttt{ck}, \texttt{ckmb}, \texttt{cl}, \texttt{cort}, \texttt{crea}, \texttt{crp}, \texttt{dbp}, \texttt{dex}, \texttt{dobu\_rate}, \texttt{dopa\_rate}, \texttt{egcs}, \texttt{eos}, \texttt{epi\_rate}, \texttt{esr}, \texttt{etco2}, \texttt{ett\_gcs}, \texttt{fgn}, \texttt{fio2}, \texttt{glu}, \texttt{hba1c}, \texttt{hbco}, \texttt{hct}, \texttt{hgb}, \texttt{hr}, \texttt{inr\_pt}, \texttt{ins}, \texttt{k}, \texttt{lact}, \texttt{lymph}, \texttt{map}, \texttt{mch}, \texttt{mchc}, \texttt{mcv}, \texttt{mech\_vent}, \texttt{methb}, \texttt{mg}, \texttt{mgcs}, \texttt{na}, \texttt{neut}, \texttt{norepi\_rate}, \texttt{o2sat}, \texttt{pco2}, \texttt{ph}, \texttt{phn\_rate}, \texttt{phos}, \texttt{plt}, \texttt{po2}, \texttt{pt}, \texttt{ptt}, \texttt{rass}, \texttt{rbc}, \texttt{rdw}, \texttt{resp}, \texttt{samp}, \texttt{sao2}, \texttt{sbp}, \texttt{spo2}, \texttt{supp\_o2}, \texttt{tco2}, \texttt{temp}, \texttt{tgcs}, \texttt{tnt}, \texttt{tri}, \texttt{urine}, \texttt{vent\_end}, \texttt{vent\_start}, \texttt{vgcs}, \texttt{wbc}.
\end{quote}

\subsection{Exact Losses and Baseline Configuration}

MAE minimizes squared reconstruction error averaged over observed feature cells at masked non-empty hours. JEPA uses elementwise squared error between predictions and layer-normalized stop-gradient target embeddings, averaged across valid target dimensions/hours. Its EMA coefficient rises from .999 to 1.0. EMA and target normalization are not evidence of collapse prevention. Core contrastive learning uses cosine similarity between normalized projected stay vectors. Each of the $2B$ view embeddings has its paired view as positive and the other non-self batch views in an NT-Xent denominator, at temperature .07. Losses differ in target space and reduction unit.

Binary mortality and AKI use two-logit cross entropy with label smoothing .1 and class weights $w_c=\sqrt{n/(2n_c)}$ from the labeled training subset. They are classification tasks, not time-to-event regression. LOS uses untransformed remaining days, squared-error training loss, and absolute-error validation/test evaluation. The constant median is the empirical training-label median evaluated on the test labels. It is not fitted on the test set.

GRU-D has hidden width 64, one recurrent layer, dropout .1, learned input/hidden decay, and the full-fine-tuning head and schedule. XGBoost receives eight summaries per variable: observed mean, SD, minimum, maximum, first, last, count, and fraction, for 672 inputs. It uses at most 1,000 trees, depth 6, learning rate .1, row/column subsampling .8, minimum child weight 1, and validation patience 20. Classification uses the native negative/positive class ratio and validation \texttt{aucpr}. Regression uses the squared-error objective and validation \texttt{mae}. These differ from neural weighting and average-precision implementation. All test AUPRC values use the shared evaluation definition. Exported per-run scientific configurations, rather than current defaults, are the reproduction reference.

\subsection{Full Baseline Matrix}

Table~\ref{tab:baseline-full} reports all full-label contextual baseline comparisons.

\begin{table*}[htbp]
\floatconts
  {tab:baseline-full}
  {\caption{Full-data contextual baselines and the best SSL mean. Classification rows report AUPRC (higher is better). LOS rows report absolute error in days (lower is better). Bold marks the best unrounded mean per row. All entries average five seeds.}}
  {\scriptsize
\begin{tabular}{llrrrrl}
\toprule
Data & Endpoint & GRU-D & XGBoost & Supervised & Best SSL & SSL method \\
\midrule
Combined & 24h mortality & \textbf{0.422} & 0.374 & 0.389 & 0.409 & MAE \\
Combined & Hospital mortality & 0.527 & 0.525 & 0.518 & \textbf{0.531} & MAE \\
Combined & AKI & 0.630 & \textbf{0.668} & 0.652 & 0.626 & JEPA \\
Combined & Remaining LOS & 2.174 & 2.340 & 2.190 & \textbf{2.118} & MAE \\
eICU-CRD & 24h mortality & \textbf{0.397} & 0.315 & 0.351 & 0.377 & MAE \\
eICU-CRD & Hospital mortality & 0.484 & 0.482 & 0.474 & \textbf{0.495} & MAE \\
eICU-CRD & AKI & 0.510 & \textbf{0.569} & 0.554 & 0.519 & MAE \\
eICU-CRD & Remaining LOS & 2.003 & 2.163 & 1.997 & \textbf{1.964} & MAE \\
MIMIC-IV & 24h mortality & \textbf{0.480} & 0.402 & 0.449 & 0.456 & Contrastive \\
MIMIC-IV & Hospital mortality & 0.577 & 0.577 & 0.565 & \textbf{0.579} & MAE \\
MIMIC-IV & AKI & 0.661 & \textbf{0.729} & 0.693 & 0.667 & MAE \\
MIMIC-IV & Remaining LOS & 2.394 & 2.613 & 2.397 & \textbf{2.359} & Contrastive \\
\bottomrule
\end{tabular}
}
\end{table*}
 
\section{Secondary Analyses}
\label{app:secondary}

\subsection{Classification Results with Seed Variation}

\tableref{tab:classification-full} gives sample SDs for the capped probe and fine-tuning, and \tableref{tab:converged-probe} (Appendix~\ref{app:converged}) gives them for the converged probe. Figure~\ref{fig:seed-differences} shows matched-seed paired differences, \MAE{} minus each comparator.

\begin{table*}[htbp]
\floatconts
  {tab:classification-full}
  {\caption{Core classification AUPRC, mean (sample SD) over five matched seeds. Probe rows use the capped linear probe. Bold marks the highest unrounded mean in each row. The supervised Transformer is defined only for full fine-tuning.}}
  {\scriptsize
\begin{tabular}{lllrrrr}
\toprule
Data & Protocol & Endpoint & MAE & JEPA & Contrastive & Supervised \\
\midrule
Combined & Fine-tune & 24h mortality & \textbf{0.409 (0.024)} & 0.391 (0.031) & 0.399 (0.029) & 0.389 (0.021) \\
Combined & Fine-tune & Hospital mortality & \textbf{0.531 (0.010)} & 0.521 (0.009) & 0.514 (0.016) & 0.518 (0.008) \\
Combined & Fine-tune & AKI & 0.626 (0.004) & 0.626 (0.005) & 0.614 (0.004) & \textbf{0.652 (0.003)} \\
Combined & Probe & 24h mortality & \textbf{0.284 (0.031)} & 0.188 (0.018) & 0.205 (0.016) & -- \\
Combined & Probe & Hospital mortality & \textbf{0.470 (0.013)} & 0.357 (0.006) & 0.392 (0.010) & -- \\
Combined & Probe & AKI & \textbf{0.447 (0.006)} & 0.327 (0.005) & 0.396 (0.005) & -- \\
eICU-CRD & Fine-tune & 24h mortality & \textbf{0.377 (0.013)} & 0.363 (0.008) & 0.360 (0.016) & 0.351 (0.022) \\
eICU-CRD & Fine-tune & Hospital mortality & \textbf{0.495 (0.005)} & 0.471 (0.007) & 0.471 (0.003) & 0.474 (0.004) \\
eICU-CRD & Fine-tune & AKI & 0.519 (0.017) & 0.515 (0.018) & 0.494 (0.012) & \textbf{0.554 (0.014)} \\
eICU-CRD & Probe & 24h mortality & \textbf{0.276 (0.015)} & 0.199 (0.017) & 0.195 (0.009) & -- \\
eICU-CRD & Probe & Hospital mortality & \textbf{0.439 (0.006)} & 0.355 (0.004) & 0.350 (0.010) & -- \\
eICU-CRD & Probe & AKI & \textbf{0.361 (0.009)} & 0.268 (0.008) & 0.286 (0.004) & -- \\
MIMIC-IV & Fine-tune & 24h mortality & 0.451 (0.045) & 0.451 (0.050) & \textbf{0.456 (0.042)} & 0.449 (0.032) \\
MIMIC-IV & Fine-tune & Hospital mortality & \textbf{0.579 (0.011)} & 0.564 (0.009) & 0.568 (0.011) & 0.565 (0.008) \\
MIMIC-IV & Fine-tune & AKI & 0.667 (0.009) & 0.642 (0.009) & 0.645 (0.009) & \textbf{0.693 (0.013)} \\
MIMIC-IV & Probe & 24h mortality & \textbf{0.312 (0.042)} & 0.236 (0.028) & 0.254 (0.035) & -- \\
MIMIC-IV & Probe & Hospital mortality & \textbf{0.510 (0.014)} & 0.415 (0.020) & 0.458 (0.023) & -- \\
MIMIC-IV & Probe & AKI & \textbf{0.504 (0.009)} & 0.410 (0.017) & 0.478 (0.009) & -- \\
\bottomrule
\end{tabular}
}
\end{table*}

\begin{figure*}[htbp]
\floatconts
  {fig:seed-differences}
  {\caption{Matched-seed classification AUPRC differences, MAE minus comparator. Positive values favor MAE. Points are mean paired differences. Bars span the five observed differences, not a confidence interval. Supervised appears only for full fine-tuning.}}
  {\includegraphics[width=.92\textwidth]{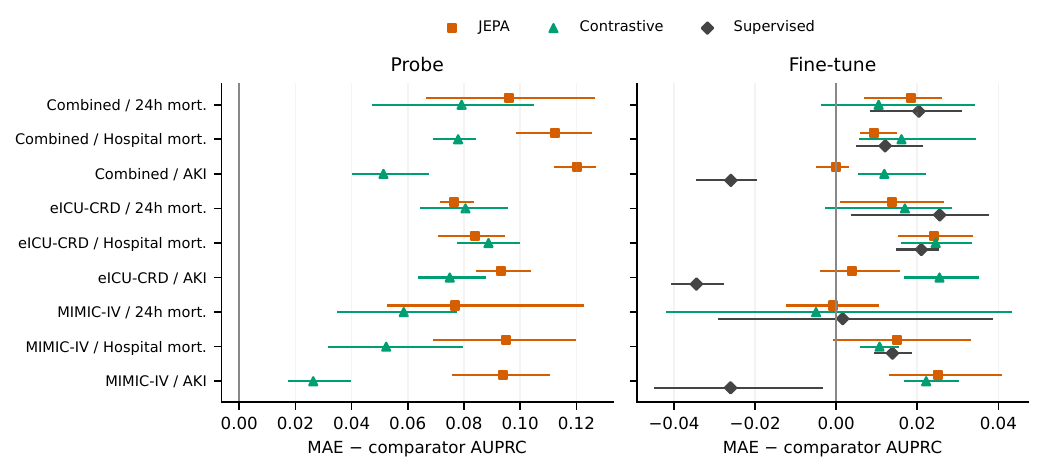}}
\end{figure*}

\subsection{Label Efficiency}
\label{app:label-efficiency}

\begin{figure*}[htbp]
\floatconts
  {fig:label-efficiency}
  {\caption{Hospital-mortality label efficiency under full fine-tuning, including the supervised Transformer at each label fraction. Points are five-seed means. Bars are sample SD across seeds. Only training labels are reduced.}}
  {\includegraphics[width=.94\textwidth]{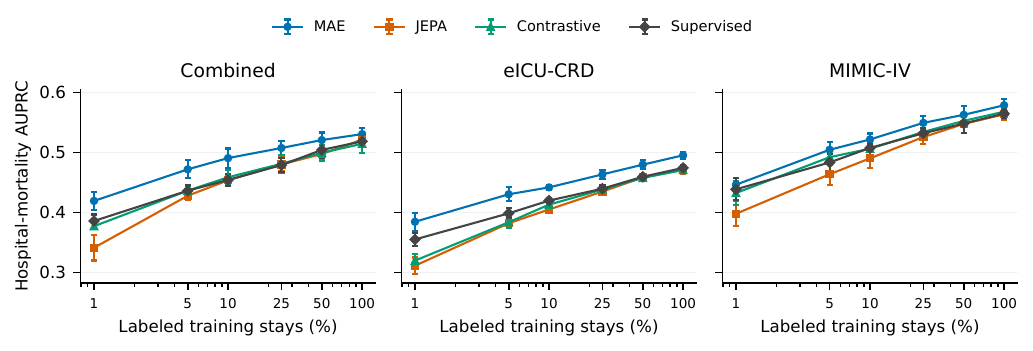}}
\end{figure*}

Only the labeled training subset is reduced. Validation and test partitions remain unchanged. Each seed draws its subsets as prefixes of one random permutation of the labeled training stays, keeping $\lfloor fn \rfloor$ of $n$ stays at fraction $f$. Subsets are therefore nested across fractions within a seed and are not class-stratified. Hospital mortality uses 1\%, 5\%, 10\%, 25\%, 50\%, and the full-label reference. Twenty-four-hour mortality starts at 5\%. AKI and LOS use 10\% and full-label references. Each cell contains five seeds.

At the lowest exported fraction for \Combined{} hospital mortality (1\%), fine-tuned AUPRC means are .419 for \MAE{}, .341 for \JEPA{}, .377 for contrastive learning, and .386 for supervised training. At 5\% labels for 24-hour mortality, the corresponding means are .277, .241, .245, and .267. When full-label references are included, MAE leads 39/45 fine-tuning cells among SSL methods, 35/45 with supervised included, and 37/39 classification probe cells among SSL methods (LOS probe cells use the superseded capped probe). Section~\ref{sec:results} instead counts only fractions below 100\%. Comparator sets and denominators must be stated together.

\begin{table*}[htbp]
\floatconts{tab:low-label}{\caption{Lowest exported label fractions, full fine-tuning. MAE and supervised means use AUPRC for classification and absolute error in days for LOS. $\Delta$ is MAE minus supervised. Positive favors MAE only for classification. SD is computed from five paired differences, not from independent error propagation.}}
{\scriptsize%
\begin{tabular}{lllrrrr}
\toprule
Data & Task & Labels & MAE & Supervised & $\Delta$ mean & $\Delta$ SD \\
\midrule
Combined & 24h mortality & 5\% & 0.277 & 0.267 & 0.010 & 0.027 \\
Combined & Hospital mortality & 1\% & 0.419 & 0.386 & 0.033 & 0.012 \\
Combined & AKI & 10\% & 0.514 & 0.507 & 0.006 & 0.021 \\
Combined & Remaining LOS & 10\% & 2.229 & 2.252 & -0.023 & 0.024 \\
eICU-CRD & 24h mortality & 5\% & 0.260 & 0.248 & 0.011 & 0.015 \\
eICU-CRD & Hospital mortality & 1\% & 0.385 & 0.355 & 0.029 & 0.019 \\
eICU-CRD & AKI & 10\% & 0.405 & 0.411 & -0.006 & 0.009 \\
eICU-CRD & Remaining LOS & 10\% & 2.067 & 2.081 & -0.014 & 0.034 \\
MIMIC-IV & 24h mortality & 5\% & 0.294 & 0.268 & 0.026 & 0.041 \\
MIMIC-IV & Hospital mortality & 1\% & 0.446 & 0.439 & 0.008 & 0.021 \\
MIMIC-IV & AKI & 10\% & 0.565 & 0.549 & 0.016 & 0.012 \\
MIMIC-IV & Remaining LOS & 10\% & 2.494 & 2.490 & 0.004 & 0.071 \\
\bottomrule
\end{tabular}
}
\end{table*}

Figure~\ref{fig:label-efficiency} shows hospital mortality. \tableref{tab:label-fraction-delta} extends the comparison to every label fraction. For hospital mortality, fine-tuned \MAE{} exceeds supervised training at every fraction in every regime. The difference is largest at low fractions on \Combined{} and \EICU{} and does not vanish at full labels. For 24-hour mortality it is also positive at every fraction on \EICU{} and \Combined{}. On \MIMIC{}, the smallest cohort, the 24-hour-mortality differences above 5\% labels change sign. For AKI, supervised training leads at full labels in all three regimes. The answer to when pretraining stops paying off therefore depends on endpoint and cohort more than on the label fraction.

Because subsets are drawn without class stratification, their positive counts vary by seed (\tableref{tab:low-label-positives}). The counts were recomputed after review from the same split and subset rules. The same rules reproduce the exported test support exactly. \MIMIC{} 24-hour mortality at 5\% labels has 31--43 training positives and also the largest AUPRC paired-difference SD in \tableref{tab:low-label}.

\begin{table*}[htbp]
\floatconts{tab:label-fraction-delta}{\caption{Mean paired difference between fine-tuned \MAE{} and the supervised Transformer at each label fraction, five seeds. Classification uses AUPRC, where positive favors \MAE{}. LOS uses absolute error in days, where negative favors \MAE{}. Dashes mark fractions that were not run.}}
{\scriptsize%
\begin{tabular}{llrrrrrr}
\toprule
Data & Task & 1\% & 5\% & 10\% & 25\% & 50\% & 100\% \\
\midrule
Combined & 24h mortality & -- & 0.010 & 0.021 & 0.033 & 0.020 & 0.020 \\
Combined & Hospital mortality & 0.033 & 0.036 & 0.036 & 0.028 & 0.016 & 0.012 \\
Combined & AKI & -- & -- & 0.006 & -- & -- & -0.026 \\
Combined & Remaining LOS & -- & -- & -0.023 & -- & -- & -0.072 \\
eICU-CRD & 24h mortality & -- & 0.011 & 0.010 & 0.026 & 0.035 & 0.026 \\
eICU-CRD & Hospital mortality & 0.029 & 0.032 & 0.022 & 0.024 & 0.020 & 0.021 \\
eICU-CRD & AKI & -- & -- & -0.006 & -- & -- & -0.034 \\
eICU-CRD & Remaining LOS & -- & -- & -0.014 & -- & -- & -0.032 \\
MIMIC-IV & 24h mortality & -- & 0.026 & -0.020 & -0.016 & -0.008 & 0.002 \\
MIMIC-IV & Hospital mortality & 0.008 & 0.021 & 0.014 & 0.017 & 0.015 & 0.014 \\
MIMIC-IV & AKI & -- & -- & 0.016 & -- & -- & -0.026 \\
MIMIC-IV & Remaining LOS & -- & -- & 0.004 & -- & -- & -0.029 \\
\bottomrule
\end{tabular}
}
\end{table*}

\begin{table*}[htbp]
\floatconts{tab:low-label-positives}{\caption{Positive training labels in each reduced-label subset, by seed, with the mean number of labeled training stays. Only the labeled training subset is reduced. Subsets are drawn per seed without class stratification. Counts were recomputed after review.}}
{\scriptsize%
\begin{tabular}{lllrrrrrrr}
\toprule
 & & & \multicolumn{6}{c}{Labeled training positives} & \\
\cmidrule(lr){4-9}
Data & Task & Labels & 42 & 123 & 456 & 789 & 1011 & Mean & Labeled stays \\
\midrule
Combined & 24h mortality & 5\% & 144 & 117 & 110 & 118 & 116 & 121.0 & 7,214.0 \\
Combined & 24h mortality & 10\% & 263 & 241 & 219 & 218 & 236 & 235.4 & 14,428.6 \\
Combined & 24h mortality & 25\% & 594 & 605 & 580 & 560 & 586 & 585.0 & 36,072.2 \\
Combined & 24h mortality & 50\% & 1,124 & 1,197 & 1,143 & 1,131 & 1,139 & 1,146.8 & 72,144.4 \\
Combined & Hospital mortality & 1\% & 144 & 164 & 150 & 159 & 150 & 153.4 & 1,442.6 \\
Combined & Hospital mortality & 5\% & 704 & 765 & 756 & 737 & 701 & 732.6 & 7,215.6 \\
Combined & Hospital mortality & 10\% & 1,427 & 1,457 & 1,489 & 1,511 & 1,444 & 1,465.6 & 14,431.8 \\
Combined & Hospital mortality & 25\% & 3,635 & 3,687 & 3,641 & 3,766 & 3,622 & 3,670.2 & 36,080.0 \\
Combined & Hospital mortality & 50\% & 7,236 & 7,366 & 7,363 & 7,508 & 7,304 & 7,355.4 & 72,160.4 \\
Combined & AKI & 10\% & 1,625 & 1,656 & 1,600 & 1,685 & 1,724 & 1,658.0 & 10,126.6 \\
eICU-CRD & 24h mortality & 5\% & 69 & 57 & 61 & 68 & 71 & 65.2 & 4,600.6 \\
eICU-CRD & 24h mortality & 10\% & 139 & 140 & 143 & 125 & 137 & 136.8 & 9,201.8 \\
eICU-CRD & 24h mortality & 25\% & 387 & 367 & 380 & 357 & 365 & 371.2 & 23,005.2 \\
eICU-CRD & 24h mortality & 50\% & 768 & 746 & 722 & 751 & 700 & 737.4 & 46,011.2 \\
eICU-CRD & Hospital mortality & 1\% & 80 & 85 & 95 & 78 & 74 & 82.4 & 919.8 \\
eICU-CRD & Hospital mortality & 5\% & 434 & 433 & 424 & 388 & 412 & 418.2 & 4,600.6 \\
eICU-CRD & Hospital mortality & 10\% & 840 & 888 & 892 & 797 & 827 & 848.8 & 9,201.8 \\
eICU-CRD & Hospital mortality & 25\% & 1,987 & 2,145 & 2,153 & 2,069 & 2,081 & 2,087.0 & 23,005.2 \\
eICU-CRD & Hospital mortality & 50\% & 4,137 & 4,234 & 4,233 & 4,217 & 4,143 & 4,192.8 & 46,011.2 \\
eICU-CRD & AKI & 10\% & 849 & 818 & 837 & 852 & 837 & 838.6 & 6,171.4 \\
MIMIC-IV & 24h mortality & 5\% & 31 & 34 & 43 & 40 & 37 & 37.0 & 2,608.6 \\
MIMIC-IV & 24h mortality & 10\% & 76 & 79 & 72 & 86 & 85 & 79.6 & 5,217.4 \\
MIMIC-IV & 24h mortality & 25\% & 216 & 199 & 190 & 207 & 202 & 202.8 & 13,044.2 \\
MIMIC-IV & 24h mortality & 50\% & 433 & 386 & 396 & 417 & 425 & 411.4 & 26,088.8 \\
MIMIC-IV & Hospital mortality & 1\% & 61 & 56 & 69 & 68 & 74 & 65.6 & 521.8 \\
MIMIC-IV & Hospital mortality & 5\% & 292 & 297 & 307 & 311 & 293 & 300.0 & 2,609.8 \\
MIMIC-IV & Hospital mortality & 10\% & 633 & 608 & 624 & 610 & 608 & 616.6 & 5,220.2 \\
MIMIC-IV & Hospital mortality & 25\% & 1,582 & 1,532 & 1,548 & 1,522 & 1,463 & 1,529.4 & 13,052.0 \\
MIMIC-IV & Hospital mortality & 50\% & 3,102 & 3,116 & 3,090 & 3,029 & 3,020 & 3,071.4 & 26,104.4 \\
MIMIC-IV & AKI & 10\% & 787 & 804 & 785 & 819 & 761 & 791.2 & 3,951.6 \\
\bottomrule
\end{tabular}
}
\end{table*}

\subsection{Cross-Database Adaptation}
\label{app:transfer}

Cross-database rows pretrain on one database and then fully fine-tune using labels from the other. They test whether a representation pretrained under one site's measurement practice can be adapted with target labels. They are not out-of-domain validation: no model is evaluated on a database without training on its labels. MAE has the best mean primary metric on three of four \MIMIC{}-to-\EICU{} tasks and ties contrastive learning on the fourth, AKI (.532 each). It leads three of four reverse-direction tasks, and contrastive learning leads reverse-direction 24-hour mortality (\tableref{tab:transfer}).

\tableref{tab:transfer-context} places cross-database \MAE{} beside \MAE{} pretrained on the target database and the supervised Transformer trained on the target. These comparators come from the core comparison with matched seeds. For mortality, cross-database encoders stay within .004--.010 AUPRC of target-database pretraining. For AKI, supervised training leads regardless of pretraining source. These are five-seed means, not tests of equivalence, and do not establish a general transfer advantage.

\begin{table*}[htbp]
\floatconts
  {tab:transfer}
  {\caption{Cross-database full-fine-tuning means over five seeds. Classification uses AUPRC. LOS uses absolute error in days. Bold marks the best mean within each row, and every mean that ties it at three decimals.}}
  {\scriptsize
\begin{tabular}{llllrrr}
\toprule
Pretrain & Target & Task & Metric & MAE & JEPA & Contrastive \\
\midrule
eICU-CRD & MIMIC-IV & 24h mortality & AUPRC & 0.447 & 0.453 & \textbf{0.454} \\
eICU-CRD & MIMIC-IV & Hospital mortality & AUPRC & \textbf{0.572} & 0.559 & 0.564 \\
eICU-CRD & MIMIC-IV & AKI & AUPRC & \textbf{0.646} & 0.643 & 0.627 \\
eICU-CRD & MIMIC-IV & Remaining LOS & Abs. error & \textbf{2.389} & 2.390 & 2.403 \\
MIMIC-IV & eICU-CRD & 24h mortality & AUPRC & \textbf{0.372} & 0.367 & 0.354 \\
MIMIC-IV & eICU-CRD & Hospital mortality & AUPRC & \textbf{0.485} & 0.477 & 0.476 \\
MIMIC-IV & eICU-CRD & AKI & AUPRC & \textbf{0.532} & 0.520 & \textbf{0.532} \\
MIMIC-IV & eICU-CRD & Remaining LOS & Abs. error & \textbf{1.966} & 1.987 & 1.998 \\
\bottomrule
\end{tabular}
}
\end{table*}

\begin{table*}[htbp]
\floatconts
  {tab:transfer-context}
  {\caption{Fine-tuned \MAE{} on each target database, by pretraining source, beside the supervised Transformer trained from scratch on the target. Means over five seeds. Classification uses AUPRC. LOS uses absolute error in days. Bold marks the best mean in each row.}}
  {\scriptsize
\begin{tabular}{lllrrr}
\toprule
Target & Task & Metric & Other-DB pretraining & Target-DB pretraining & Supervised \\
\midrule
eICU-CRD & 24h mortality & AUPRC & 0.372 & \textbf{0.377} & 0.351 \\
eICU-CRD & Hospital mortality & AUPRC & 0.485 & \textbf{0.495} & 0.474 \\
eICU-CRD & AKI & AUPRC & 0.532 & 0.519 & \textbf{0.554} \\
eICU-CRD & Remaining LOS & Abs. error & 1.966 & \textbf{1.964} & 1.997 \\
MIMIC-IV & 24h mortality & AUPRC & 0.447 & \textbf{0.451} & 0.449 \\
MIMIC-IV & Hospital mortality & AUPRC & 0.572 & \textbf{0.579} & 0.565 \\
MIMIC-IV & AKI & AUPRC & 0.646 & 0.667 & \textbf{0.693} \\
MIMIC-IV & Remaining LOS & Abs. error & 2.389 & \textbf{2.368} & 2.397 \\
\bottomrule
\end{tabular}
}
\end{table*}

\subsection{Encoder Capacity and Subgroup Auditing}
\label{app:capacity-subgroups}

The capacity analysis compares MAE and supervised Transformers on \MIMIC{} 24-hour mortality at 5\%, 10\%, and 50\% labels. Under full fine-tuning, medium and large MAE models exceed their corresponding supervised models by .013--.028 mean AUPRC across the six size--fraction cells, each with five seeds. This one-task, one-database slice does not test larger JEPA or contrastive encoders and cannot establish a general scaling rule.

Subgroup auditing records maximum-minus-minimum performance gaps across eligible gender, age, and race groups, requiring at least 50 unique patients and a defined metric per group. In the core full-label classification cells, across-seed mean AUPRC gaps range from .004 to .095 for gender, .033 to .172 for age, and .080 to .513 for race. These ranges cover 63 gender cells, 63 age cells, and 42 race cells. Race metrics are unavailable for \EICU{}, and \Combined{} race analyses use \MIMIC{}-origin records. A 50-patient threshold does not ensure enough positive outcomes for stable AUPRC. These summaries describe gaps without ranking methods by fairness or attributing their causes. They do not assess intersectional groups or subgroup calibration.

\subsection{TS2Vec Context}

\label{app:ts2vec}

Our TS2Vec-style extension retains the SLICES Transformer and value--mask interface, rather than the original dilated convolutional encoder. It adds Gaussian noise of SD .01 to observed values, independently masks half the hours in each view, and uses no temporal crop in the reference configuration. A width-256/output-64 projection precedes temperature-.05 NT-Xent over overlapping timestep pairs. Negatives combine other times and other instances in one denominator. This differs from original TS2Vec's separate temporal and instance losses. Four max-pooling scales (1, 2, 4, 8 hours) use mask-aware validity and an average over active scales. Thus these results characterize this extension, not original TS2Vec's performance \citep{yue2022ts2vec}.

\begin{table*}[htbp]
\floatconts
  {tab:ts2vec}
  {\caption{TS2Vec-style extension versus the core contrastive implementation. Entries count dataset--task cells with the better unrounded mean among 12 cells per protocol.}}
  {%
\begin{tabular}{lrr}
\toprule
Protocol & TS2Vec-style & Core contrastive \\
\midrule
Fine-tune & 8 & 4 \\
Probe & 0 & 12 \\
\bottomrule
\end{tabular}
}
\end{table*}

\tableref{tab:ts2vec} counts the cells each implementation wins. The full-fine-tuning split is task dependent. For example, TS2Vec is higher on \MIMIC{} 24-hour mortality (.465 versus .456), while core contrastive is lower on \MIMIC{} LOS (2.359 versus 2.370). Under the capped probe, the core contrastive mean is better in every cell, including \MIMIC{} AKI (.478 versus .441) and LOS (2.592 versus 2.722). The TS2Vec-style encoders were not refit with converged or loss-matched probes. These five-seed differences do not establish equivalence or general dominance.
 
\subsection{Demonstrated Extension Workflow}
\label{app:extension-workflow}

\begin{figure*}[htbp]
\floatconts
  {fig:benchmark}
  {\caption{\SLICES{} benchmark. The reusable path from fixed specification through objective extension and encoder transfer to auditable results. Comparison checks enforce declared settings and export coverage. They do not prove that a training run used its declared private data. Objective-specific modules and compute remain different.}}
  {\includegraphics[width=\textwidth]{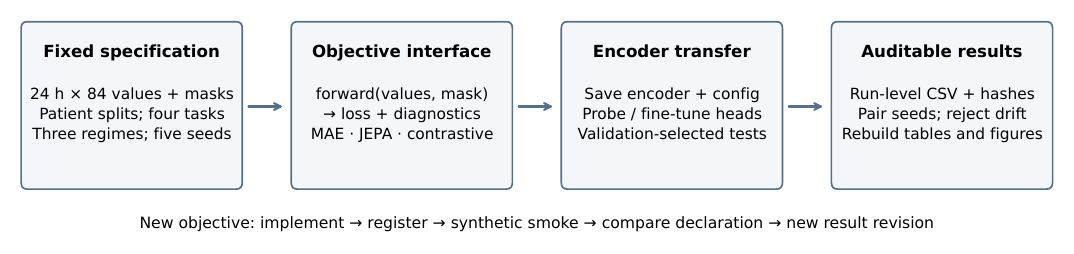}}
\end{figure*}

The code repository's \texttt{TS2VEC\_CASE.md} gives a runnable path through the existing extension. Objective code, its configuration dataclass, registry entries, encoder-compatibility entry, and SSL YAML define the replaceable components. Clinical preprocessing, feature order, the shared encoder, final-epoch transfer, task heads, protocol schedules, primary metrics, and result identities define the retained scientific contract. The case adds execution and reporting code. It does not alter the objective or the archived results.

The command \texttt{extension.py --out extension-output} uses 80 artificial patients, the shared 70/15/15 split function, train-only normalization, and the full 159,808-parameter encoder. It runs two SSL epochs and two epochs for each synthetic binary/regression task under each protocol. These reduced budgets, artificial targets, and single seed are recorded as departures from the scientific contract. Checks cover exact v3 encoder transfer, discarded projection weights, unchanged frozen parameters, updated fine-tuned parameters, validation selection, and exact test-checkpoint reload. Synthetic metrics remain in a separate output file. The report on archived runs instead joins 120 existing TS2Vec runs to their matched core-contrastive runs, retaining means, sample SDs, paired differences, and configuration IDs.

Validation records distinguish implementation assistance and failed attempts from successful execution of the extracted archive's instructions. This is a development-agent check on a fresh environment, with no independent human reuse study. It does not test credentialed extraction, clinical label construction, or historical numerical retraining equivalence, which Appendix~\ref{app:reproduction} examines for a \MIMIC{} slice.

\subsection{Post-Hoc Random-Encoder Diagnostic}
\label{app:random}

The random diagnostic contains 40 task--dataset--seed rows over \MIMIC{} and \EICU{}. It freezes an untrained instance of the same default Transformer and trains only the linear task head from cached embeddings. Pairing those rows with the three core SSL probes gives 120 comparisons. The random probe is worse in 118. The exceptions are both against \JEPA{} on \EICU{}, by at most .001 AUPRC, for AKI and 24-hour mortality in one run each. This analysis was produced after the archived results and was not part of the core family. The raw/mask control below is another separate diagnostic.

\subsection{JEPA, Checkpoint, and Hyperparameter Diagnostics}
\label{app:diagnostics}

The separate JEPA diagnostic covers all 15 core pretraining runs. Across logged steps, the minimum target representation SD is .37 and the maximum mean pairwise cosine is .82. The mean final SD is 1.17. These statistics cannot rule out partial collapse or establish downstream quality. The historical checkpoint audit covers 45 core pretraining encoders. Core downstream runs use the final 100-epoch encoder. Retained JEPA validation-checkpoint files are from epochs 3--5. These are not final encoders and do not mean training ended early.

The checkpoint diagnostic completes 60 frozen probes, or 30 final-versus-best-SSL-validation pairs, over the three core objectives, five seeds, and MIMIC-IV hospital mortality and LOS. Both arms use the same cached CPU implementation, patient splits, train-only normalization, head initialization and batch RNG. Linear heads use the reference AdamW/cosine settings, validation patience and epoch cap. LOS trains with squared error and selects/reports absolute error. Validation alone selects the head, with the earliest epoch retained on a tie. The test set is evaluated once afterward. Producer checks matched all 30 encoder hashes to the archived encoder audit. The public audit reconstructs paired statistics and all 60 selections from saved validation histories.

\begin{table*}[htbp]
\floatconts{tab:checkpoint-sensitivity}
{\caption{Separate MIMIC-IV cached CPU probe diagnostic. Mean $\pm$ sample SD over five seeds. Mortality uses AUPRC. LOS uses absolute error in days. $\Delta$ is best-SSL-validation minus rerun-final. Brackets give the observed paired range. These final-arm values are reruns, not substituted historical scores.}}
{\scriptsize%
\begin{tabular}{llrlll}
\toprule
Task & Encoder & $n$ & Rerun final & Best SSL val. & $\Delta$ mean [range] \\
\midrule
Hospital mortality & MAE & 5 & 0.5093 $\pm$ 0.0135 & 0.5096 $\pm$ 0.0139 & +0.0003 [-0.0009, +0.0019] \\
Hospital mortality & JEPA & 5 & 0.4139 $\pm$ 0.0196 & 0.3733 $\pm$ 0.0094 & -0.0406 [-0.0563, -0.0166] \\
Hospital mortality & Contrastive & 5 & 0.4574 $\pm$ 0.0236 & 0.4573 $\pm$ 0.0236 & -0.0001 [-0.0003, +0.0000] \\
Remaining LOS & MAE & 5 & 2.7655 $\pm$ 0.0651 & 2.7644 $\pm$ 0.0628 & -0.0011 [-0.0047, +0.0019] \\
Remaining LOS & JEPA & 5 & 2.6758 $\pm$ 0.0903 & 2.9396 $\pm$ 0.0916 & +0.2638 [+0.1909, +0.3689] \\
Remaining LOS & Contrastive & 5 & 2.5946 $\pm$ 0.0662 & 2.5949 $\pm$ 0.0655 & +0.0003 [-0.0009, +0.0018] \\
\bottomrule
\end{tabular}
}
\end{table*}

JEPA worsens in all five seeds on both tasks (Table~\ref{tab:checkpoint-sensitivity}). For MAE and contrastive learning, the choice between final and best-validation encoders changes mean AUPRC and mean LOS error by at most .001 (AUPRC or days). MAE leads mortality and contrastive leads LOS under both policies in every seed, while JEPA's LOS mean rank drops from second to third. These LOS ranks use the capped squared-error probe, whose ordering did not survive loss matching (Section~\ref{sec:results}). The per-seed selections, validation histories, and paired differences are in \texttt{results/checkpoint-sensitivity/}, and \texttt{CHECKPOINT\_CASE.md} documents how to recompute them. Rerun-final cell means differ from historical means by at most .001 AUPRC and .01 LOS days. The largest absolute per-seed LOS deviation is .09 days. Cached CPU execution and dedicated paired RNG differ from historical GPU/Lightning execution. This diagnostic took about 20 wall minutes on CPU, with no failed or timed-out workers. It supports scoped checkpoint sensitivity, not equivalence, general rank stability, or conclusions about other databases or full fine-tuning.

Table~\ref{tab:hp} contains only MIMIC-IV 24-hour mortality, full fine-tuning, full labels, and five seeds per level. Learning-rate values are upstream pretraining rates. Mask sweeps change their respective objective's mask ratio. The largest within-sweep range of mean AUPRC is .018. This targeted slice establishes neither hyperparameter insensitivity nor objective-specific optimal tuning. Its test-set rankings are not used to select a new headline configuration.

\begin{table*}[htbp]
\floatconts{tab:hp}{\caption{Targeted pretraining hyperparameter diagnostics. Downstream AUPRC mean and sample SD, $n=5$ per level.}}
{\scriptsize%
\begin{tabular}{lllrr}
\toprule
Sweep & Objective & Level & AUPRC & SD \\
\midrule
lr sensitivity & Contrastive & 0.0002 & 0.462 & 0.035 \\
lr sensitivity & Contrastive & 0.0005 & 0.465 & 0.033 \\
lr sensitivity & Contrastive & 0.002 & 0.459 & 0.042 \\
lr sensitivity & JEPA & 0.0002 & 0.467 & 0.028 \\
lr sensitivity & JEPA & 0.0005 & 0.461 & 0.049 \\
lr sensitivity & JEPA & 0.002 & 0.449 & 0.053 \\
lr sensitivity & MAE & 0.0002 & 0.462 & 0.047 \\
lr sensitivity & MAE & 0.0005 & 0.451 & 0.051 \\
lr sensitivity & MAE & 0.002 & 0.457 & 0.050 \\
mask ratio sensitivity & JEPA & 0.3 & 0.454 & 0.057 \\
mask ratio sensitivity & JEPA & 0.75 & 0.447 & 0.029 \\
mask ratio sensitivity & MAE & 0.3 & 0.464 & 0.033 \\
mask ratio sensitivity & MAE & 0.75 & 0.447 & 0.035 \\
view mask sensitivity & Contrastive & 0.3 & 0.460 & 0.035 \\
view mask sensitivity & Contrastive & 0.75 & 0.447 & 0.061 \\
\bottomrule
\end{tabular}
}
\end{table*}

\subsection{Raw-Input and Mask-Only CPU Control}
\label{app:raw}

This post-hoc diagnostic uses only MIMIC-IV hospital mortality and the five canonical patient splits. A linear logistic classifier receives either 2,016 flattened mask entries or those entries plus 2,016 normalized values. Normalization uses the same full-cohort training partition. Task labels are filtered after patient assignment. Fixed SGD uses logistic loss, $L_2$ coefficient $10^{-4}$, the library's optimal learning-rate schedule, square-root-balanced sample weights, and 20 epochs. The epoch with highest validation average precision is evaluated once on the test set. No label smoothing or hyperparameter search is used. The implementation and input-file SHA256 values are recorded separately.

Test AUPRC is .301 (.025) for masks and .404 (.013) for values plus masks, mean (sample SD). Mean test support is 11,157 stays and 1,310 deaths. Total wall time is 136 seconds on a CPU limited to two threads, including both representations and all five seeds, under a 900-second hard timeout. Every mask-only run selects epoch 20, so further optimization could improve this control. The result does not establish the performance of an optimized raw-input model, nor does it isolate SSL gains from differences in optimization, regularization, input dimension, or head training.
 
\clearpage
\section{Post-Review Diagnostics}
\label{app:post-review}

The analyses in this section were run after review, to answer reviewer questions. They are stored separately from the archived results, with their scripts and checksums, in \texttt{results/ml4h-rebuttal-v1/}. \texttt{render\_rebuttal.py} regenerates their tables from those aggregates.

\subsection{Converged and Loss-Matched Probes}
\label{app:converged}

The classification probe is an L2-regularized logistic regression fitted with L-BFGS to convergence, with the reference square-root-balanced class weights and 17 log-spaced regularization strengths selected on validation AUPRC. The absolute-error LOS probe uses a Huber loss with a .01-day threshold and is selected on validation absolute error. The ridge probe is selected on validation squared error. Each selected probe scores the test set once. Embeddings come from the archived final-epoch encoders, whose SHA256 digests match the encoder audit in 45/45 cases. Training rows, positives, and test rows match the archived results in every cell. In three checked cells, cached embeddings match the benchmark forward pass to within $9.5\times10^{-7}$. Embeddings are standardized with training statistics. The converged classification probe differs from the capped probe in optimizer (L-BFGS to convergence instead of AdamW for at most 50 epochs), regularization (tuned L2 instead of fixed weight decay), input scaling, and label smoothing (none instead of .1). Class weights are the same. All logistic, ridge, and Huber fits met their convergence criterion. No selected strength lies at the strongest end of its grid, and the selections at the weakest end (6/135 logistic, 15/45 ridge, 21/45 Huber) lie on a flat part of the regularization path.

\tableref{tab:converged-probe} gives the converged-probe results with seed variation and paired advantages. AUROC means and objective rankings for both probes are in \texttt{p1\_objective\_ranking.csv}.

\begin{table*}[htbp]
\floatconts
  {tab:converged-probe}
  {\caption{Converged frozen probe. Test AUPRC, mean (sample SD) over five seeds, on the same frozen embeddings as the capped probe. $\Delta$ is the mean paired difference between \MAE{} and the stronger alternative, for the converged (conv.) and capped (cap.) probe. Bold marks the highest mean.}}
  {\scriptsize
\begin{tabular}{llrrrrr}
\toprule
Data & Endpoint & MAE & JEPA & Contrastive & $\Delta$ conv. & $\Delta$ cap. \\
\midrule
Combined & 24h mortality & \textbf{0.309 (0.033)} & 0.208 (0.012) & 0.239 (0.023) & 0.070 & 0.079 \\
Combined & Hospital mortality & \textbf{0.470 (0.012)} & 0.363 (0.006) & 0.395 (0.010) & 0.075 & 0.078 \\
Combined & AKI & \textbf{0.451 (0.005)} & 0.332 (0.004) & 0.402 (0.006) & 0.049 & 0.051 \\
eICU-CRD & 24h mortality & \textbf{0.287 (0.017)} & 0.219 (0.019) & 0.215 (0.003) & 0.068 & 0.076 \\
eICU-CRD & Hospital mortality & \textbf{0.439 (0.007)} & 0.359 (0.005) & 0.356 (0.009) & 0.081 & 0.084 \\
eICU-CRD & AKI & \textbf{0.366 (0.010)} & 0.275 (0.004) & 0.293 (0.006) & 0.074 & 0.075 \\
MIMIC-IV & 24h mortality & \textbf{0.357 (0.035)} & 0.298 (0.018) & 0.322 (0.037) & 0.035 & 0.059 \\
MIMIC-IV & Hospital mortality & \textbf{0.520 (0.008)} & 0.440 (0.016) & 0.471 (0.018) & 0.049 & 0.052 \\
MIMIC-IV & AKI & \textbf{0.510 (0.008)} & 0.433 (0.017) & 0.492 (0.011) & 0.018 & 0.026 \\
\bottomrule
\end{tabular}
}
\end{table*}

Rerunning the capped recipe on fixed embeddings with five head-initialization seeds shows that its result is not a stable function of the embedding for 24-hour mortality and LOS, where early stopping on a noisy validation metric decides the outcome. \tableref{tab:los-reference} gives the capped-probe LOS results that \tableref{tab:los} replaces. The Huber probe uses a fixed .01-day threshold rather than an exact absolute-error loss.

\begin{table*}[htbp]
\floatconts
  {tab:los-reference}
  {\caption{Capped-probe LOS results from the archived results, superseded for probing by \tableref{tab:los}. Absolute error in days, mean (sample SD), $n=5$. Probes train with squared error for at most 50 epochs and are selected on validation absolute error. The constant predicts each seed's training-label median. Duplicate logs are collapsed by dataset and seed. Bold marks the lowest model mean within a protocol.}}
  {\scriptsize
\begin{tabular}{llrrrrr}
\toprule
Data & Protocol & MAE & JEPA & Contrastive & Supervised & Constant median \\
\midrule
Combined & Fine-tune & \textbf{2.118 (0.029)} & 2.135 (0.013) & 2.152 (0.024) & 2.190 (0.031) & 2.403 (0.025) \\
Combined & Probe & 2.425 (0.027) & 2.475 (0.023) & \textbf{2.309 (0.029)} & -- & 2.403 (0.025) \\
eICU-CRD & Fine-tune & \textbf{1.964 (0.017)} & 1.977 (0.034) & 2.008 (0.031) & 1.997 (0.030) & 2.176 (0.006) \\
eICU-CRD & Probe & 2.191 (0.011) & 2.249 (0.022) & \textbf{2.139 (0.011)} & -- & 2.176 (0.006) \\
MIMIC-IV & Fine-tune & 2.368 (0.051) & 2.371 (0.069) & \textbf{2.359 (0.069)} & 2.397 (0.066) & 2.733 (0.074) \\
MIMIC-IV & Probe & 2.755 (0.058) & 2.679 (0.070) & \textbf{2.592 (0.076)} & -- & 2.733 (0.074) \\
\bottomrule
\end{tabular}
}
\end{table*}

\subsection{Combined Results by Source Database}
\label{app:per-source}

\tableref{tab:per-source} scores the \Combined{}-regime models separately on the \MIMIC{} and \EICU{} portions of the \Combined{} test split. Probe columns use the converged probe. Fine-tune and supervised columns rerun test inference on the archived reference checkpoints, after verifying their SHA256 digests. Pooled metrics from this inference reproduce the archived results to within $2.4\times10^{-5}$. Prevalence differs between portions, so AUPRC should be compared within a portion. AUROC, which is higher on the \MIMIC{} portion in all 30 objective--protocol--task cells, is in the two \texttt{p3\_*\_per\_source\_aggregated.csv} files. For LOS, the converged absolute-error probe puts \MAE{} first on the \EICU{} portion and last on the \MIMIC{} portion, within .021 days there. Fine-tuned \MAE{} has the lowest LOS error on both portions.

\begin{table*}[htbp]
\floatconts
  {tab:per-source}
  {\caption{\Combined{}-regime models scored separately on the two source portions of the \Combined{} test split. Means over five seeds. Classification uses AUPRC. LOS uses absolute error in days, with the Huber probe in the probe columns. Prev.\ is the mean test prevalence of the portion. Bold marks the best mean within each protocol.}}
  {\scriptsize
\begin{tabular}{llrrrrrrrr}
\toprule
Task & Portion & Prev. & \multicolumn{3}{c}{Probe (converged)} & \multicolumn{4}{c}{Fine-tune} \\
\cmidrule(lr){4-6} \cmidrule(lr){7-10}
 & & & MAE & JEPA & Con. & MAE & JEPA & Con. & Sup. \\
\midrule
24h mortality & eICU-CRD & 0.016 & \textbf{0.302} & 0.195 & 0.225 & \textbf{0.392} & 0.377 & 0.384 & 0.373 \\
24h mortality & MIMIC-IV & 0.016 & \textbf{0.334} & 0.250 & 0.276 & \textbf{0.448} & 0.423 & 0.429 & 0.424 \\
Hospital mortality & eICU-CRD & 0.093 & \textbf{0.443} & 0.337 & 0.362 & \textbf{0.498} & 0.485 & 0.479 & 0.484 \\
Hospital mortality & MIMIC-IV & 0.118 & \textbf{0.507} & 0.401 & 0.442 & \textbf{0.576} & 0.570 & 0.563 & 0.566 \\
AKI & eICU-CRD & 0.139 & \textbf{0.373} & 0.261 & 0.309 & 0.553 & 0.556 & 0.541 & \textbf{0.582} \\
AKI & MIMIC-IV & 0.204 & \textbf{0.525} & 0.396 & 0.486 & 0.697 & 0.695 & 0.685 & \textbf{0.720} \\
\midrule
Remaining LOS & eICU-CRD & -- & \textbf{2.013} & 2.049 & 2.036 & \textbf{1.967} & 1.982 & 2.001 & 2.034 \\
Remaining LOS & MIMIC-IV & -- & 2.513 & 2.509 & \textbf{2.493} & \textbf{2.387} & 2.408 & 2.422 & 2.468 \\
\bottomrule
\end{tabular}
}
\end{table*}

\subsection{Compute}
\label{app:compute}

All 45 core pretraining runs completed the 100-epoch schedule at batch size 256 in fp32 on one NVIDIA L4. The reference runs shared the GPU, about four at a time, so their wall-clock times in \tableref{tab:compute} are not exclusive GPU time. One \MAE{} \MIMIC{} run that ran nearly alone took 0.92 hours, against about 2.25 hours for each of the four that ran concurrently. We therefore also timed each objective alone on an idle L4 of the same machine type, with the reference configuration on real \MIMIC{} batches, 20 warm-up steps, and 300 timed training steps. Projected hours cover training steps only, without validation or checkpointing. Step time did not depend on the database and changed by at most 3\% with eight data-loader workers. In a monitored repeat, GPU utilization peaked at 13--21\% (typically 10--18\%) while one host CPU core was saturated, so wall time does not follow FLOPs. The cost order depends on the measure. FLOPs and shared-GPU wall-clock rank \MAE{} $<$ contrastive $<$ \JEPA{}, while the idle GPU ranks contrastive $<$ \MAE{} $<$ \JEPA{}. Dividing each run's shared wall-clock time by its mean number of concurrent runs gives about 55 L4-hours for the 45 runs. Projecting the idle-GPU step times onto their optimizer steps gives about 63 hours of training steps alone. Under an equal-FLOPs budget, \JEPA{} would receive about 45 \MAE{}-equivalent epochs and contrastive learning about 76. In the checkpoint diagnostic, early \JEPA{} checkpoints of the 100-epoch schedule probed worse, but a shorter full schedule was not tested, nor was longer contrastive training.

\begin{table*}[htbp]
\floatconts
  {tab:compute}
  {\caption{Pretraining cost per objective. FLOPs are forward FLOPs per sample. Wall-clock hours are W\&B runtimes on a shared L4, mean (sample SD) over five seeds. Idle-GPU columns time training steps alone on an idle L4 and project them to the 100-epoch schedule. The \JEPA{} EMA target copy is not trainable.}}
  {\scriptsize
\setlength{\tabcolsep}{4pt}
\begin{tabular}{@{}lrrrrrrrrr@{}}
\toprule
 & Fwd. MFLOPs & Trainable & Idle L4 & \multicolumn{2}{c}{MIMIC-IV (h)} & \multicolumn{2}{c}{eICU-CRD (h)} & \multicolumn{2}{c}{Combined (h)} \\
\cmidrule(lr){5-6} \cmidrule(lr){7-8} \cmidrule(lr){9-10}
Objective & per sample & params & ms/step & Shared & Idle & Shared & Idle & Shared & Idle \\
\midrule
MAE & 4.78 & 269,460 & 119.6 & 1.99 (0.60) & 0.68 & 3.94 (0.04) & 1.19 & 6.16 (0.04) & 1.89 \\
JEPA & 10.60 & 189,440 + 159,808 EMA & 179.7 & 3.00 (0.02) & 1.02 & 5.34 (0.03) & 1.81 & 8.32 (0.06) & 2.86 \\
Contrastive & 6.32 & 193,408 & 97.6 & 2.67 (0.01) & 0.56 & 4.76 (0.05) & 0.99 & 7.41 (0.03) & 1.56 \\
\bottomrule
\end{tabular}
}
\end{table*}

\subsection{Reproduction with the Maintained Framework}
\label{app:reproduction}

The reference runs used the historical implementation. We reran a \MIMIC{} slice with the maintained repository code, starting from the processed tensors (\tableref{tab:reproduction}). Training configurations equal the reference apart from data path, data-loader workers, and device. Downstream runs from the archived reference encoders (seed 42) isolate the downstream code. Full-pipeline runs (seeds 42 and 123) also retrain the encoders. All six pretraining runs completed the 100-epoch schedule with the reference number of optimizer steps, and their final validation losses lie within 0.63 across-seed SDs of the reference. The reruns used a laptop CPU, while the reference used an L4. CPU and GPU draw different random streams for masks and dropout, so a rerun with the same seed is a different realization of the same configuration. The archived-encoder probes are deterministic and match to within $7.6\times10^{-6}$. In the full pipeline, the objective order differs from the reference in three fine-tuning cells whose five-seed reference means are near-ties. LOS fine-tuning error is higher than the reference in all nine comparisons. Twenty of the 24 seed-123 reference downstream runs, and 37 of the 420 core downstream runs overall, had been resumed from their last checkpoint after an interruption, while the reruns ran uninterrupted. This slice cannot separate chance, resumption, and a small CPU-versus-GPU effect.

\begin{table*}[htbp]
\floatconts
  {tab:reproduction}
  {\caption{Maintained-framework reruns against the same-seed reference runs on \MIMIC{}. $|\Delta|$ is the absolute difference in the test metric. SD columns count runs whose primary-metric difference lies within one or two reference across-seed SDs of the same objective, task, and protocol.}}
  {\scriptsize
\begin{tabular}{llrrrrr}
\toprule
Starting point & Protocol & Runs & Mean $|\Delta|$ AUPRC (max) & Mean $|\Delta|$ LOS days (max) & $\leq$1 SD & $\leq$2 SD \\
\midrule
Archived encoders & Probe & 12 & $<10^{-5}$ & 0 (0) & 12 & 12 \\
Archived encoders & Fine-tune & 12 & 0.010 (0.040) & 0.029 (0.045) & 12 & 12 \\
Full pipeline & Probe & 24 & 0.008 (0.031) & 0.007 (0.016) & 23 & 24 \\
Full pipeline & Fine-tune & 24 & 0.010 (0.048) & 0.044 (0.074) & 19 & 24 \\
\bottomrule
\end{tabular}
}
\end{table*}
 
\clearpage
\section{Evidence and Reproducibility Boundary}
\label{app:reproducibility}

The archived evidence bundle used for the paper contains:
\begin{itemize}
  \item an aggregate result table with one row per scientific cell and seed summaries.
  \item a per-seed table used for matched differences and coverage checks.
  \item long-form metrics and retained diagnostic test outputs.
  \item label-efficiency, classical-context, and TS2Vec comparison tables.
  \item an evidence manifest recording row counts, schemas, analysis revision, and source hashes.
\end{itemize}

The core audit filter is: core SSL analysis, default model size, full label fraction, the three data regimes, four tasks, and the two adaptation protocols. It yields 84 aggregate rows with five seeds each and 420 per-seed rows. The random-encoder diagnostic is stored separately and is not merged into the archived core ranking. No patient-level row is included in the paper source or code repository.

The code repository also includes \texttt{thesis-coverage-evidence.json}, with source hashes and aggregate cells for the capacity, subgroup, and LOS secondary-metric summaries. Post-review diagnostics (Appendix~\ref{app:post-review}) and the recomputed reduced-label positive counts (\texttt{results/camera-ready-v1/}) are separate packages with their own provenance notes. They do not modify the archived results. The paper source includes LaTeX, generated table fragments and vector figures, a bibliography, and the official class/style files. It builds with PDF\LaTeX{} and BibTeX through \texttt{latexmk}.
 
\end{document}